\documentclass[10pt,twocolumn,letterpaper]{article}

\usepackage[pagenumbers]{wacv} % Combined arXiv version with continuous page numbers
\usepackage{microtype}
\usepackage{dblfloatfix}
\usepackage[accsupp]{axessibility}

\usepackage{placeins}
\usepackage{bm}

\providecommand{\best}[1]{%
    \ensuremath{\bm{#1}}%
}

\providecommand{\secondbest}[1]{%
    \underline{\ensuremath{#1}}%
}

\providecommand{\resultpm}[2]{%
    \ensuremath{#1_{\scriptscriptstyle \pm #2}}%
}

\providecommand{\bestpm}[2]{%
    \ensuremath{\bm{#1}_{\scriptscriptstyle \pm #2}}%
}

\providecommand{\secondbestpm}[2]{%
    \underline{\ensuremath{#1_{\scriptscriptstyle \pm #2}}}%
}

\definecolor{wacvblue}{rgb}{0.21,0.49,0.74}
\usepackage[pagebackref,breaklinks,colorlinks,allcolors=wacvblue]{hyperref}

\usepackage{graphicx}
\usepackage{booktabs}
\usepackage{multirow}
\usepackage{amsmath}
\usepackage{bm}

\providecommand{\best}[1]{%
    \ensuremath{\bm{#1}}%
}

\providecommand{\secondbest}[1]{%
    \underline{\ensuremath{#1}}%
}

\providecommand{\resultpm}[2]{%
    \ensuremath{#1_{\scriptscriptstyle \pm #2}}%
}

\providecommand{\bestpm}[2]{%
    \ensuremath{\bm{#1}_{\scriptscriptstyle \pm #2}}%
}

\providecommand{\secondbestpm}[2]{%
    \underline{\ensuremath{#1_{\scriptscriptstyle \pm #2}}}%
}

\def\wacvPaperID{765} % *** Enter the WACV Paper ID here
\def\confName{WACV}
\def\confYear{2027}

\title{Flownomaly: Single-Step Conditional-Mean Recovery for Few-Shot Anomaly Detection}

\author{Bernardo Perrone Ribeiro \qquad Erkut Akdag \qquad Egor Bondarev\\
AIMS Group, Department of Electrical Engineering, Eindhoven University of Technology\\
{\tt\small \{b.perrone.de.menezes.bulcao.ribeiro, e.akdag, e.bondarev\}@tue.nl}
}

\begin{document}
\maketitle
\begin{abstract}
Few-shot industrial anomaly detection requires modeling category-specific normality from only a handful of defect-free images. Common approaches tie inference to stored exemplars, impose rigid linear structure, or risk reproducing anomalous content. \textbf{Flownomaly} instead casts the problem as normal-counterpart recovery in frozen DINOv2 feature space. Given normal supports, a velocity field is learned through conditional flow matching over frozen patch tokens to transport Gaussian noise toward normal features. Along straight paths, a single recovery step using one evaluation of the optimal field yields the posterior mean of the clean normal token without trajectory integration. The operating time governs how strongly recovery modifies a query token. With no anomalous validation samples available for tuning, a normal-only Jacobian gain-to-noise criterion selects this time from the supports alone. Squared residuals between query tokens and their recovered counterparts yield anomaly signals for both dense localization and image-level detection. Across 1-, 2-, and 4-shot benchmarks on MVTec-AD and VisA, together with sample-shot evaluations on Real-IAD, Flownomaly achieves state-of-the-art performance against recent category-specific methods. In the 1-shot setting, image- and pixel-level AUROC reach 97.8\% and 98.0\% on MVTec-AD and 95.3\% and 98.7\% on VisA, with a 2.1-point image-level AUROC improvement over the strongest baseline on VisA.
Code is available at \url{https://github.com/b-rbmp/Flownomaly}.
\end{abstract}    
\section{Introduction}
\label{sec:intro}

Industrial visual anomaly detection aims to identify rare and unpredictable defects, including scratches, contamination, misplaced components, and structural damage~\cite{mvtec}. Few-shot anomaly detection (FSAD) addresses this setting by characterizing category-specific normality from only a handful of defect-free reference images and detecting unseen defects at both image and pixel levels~\cite{huang2022registration,winclip}.

\begin{figure}[t]
\centering
\includegraphics[width=\linewidth]{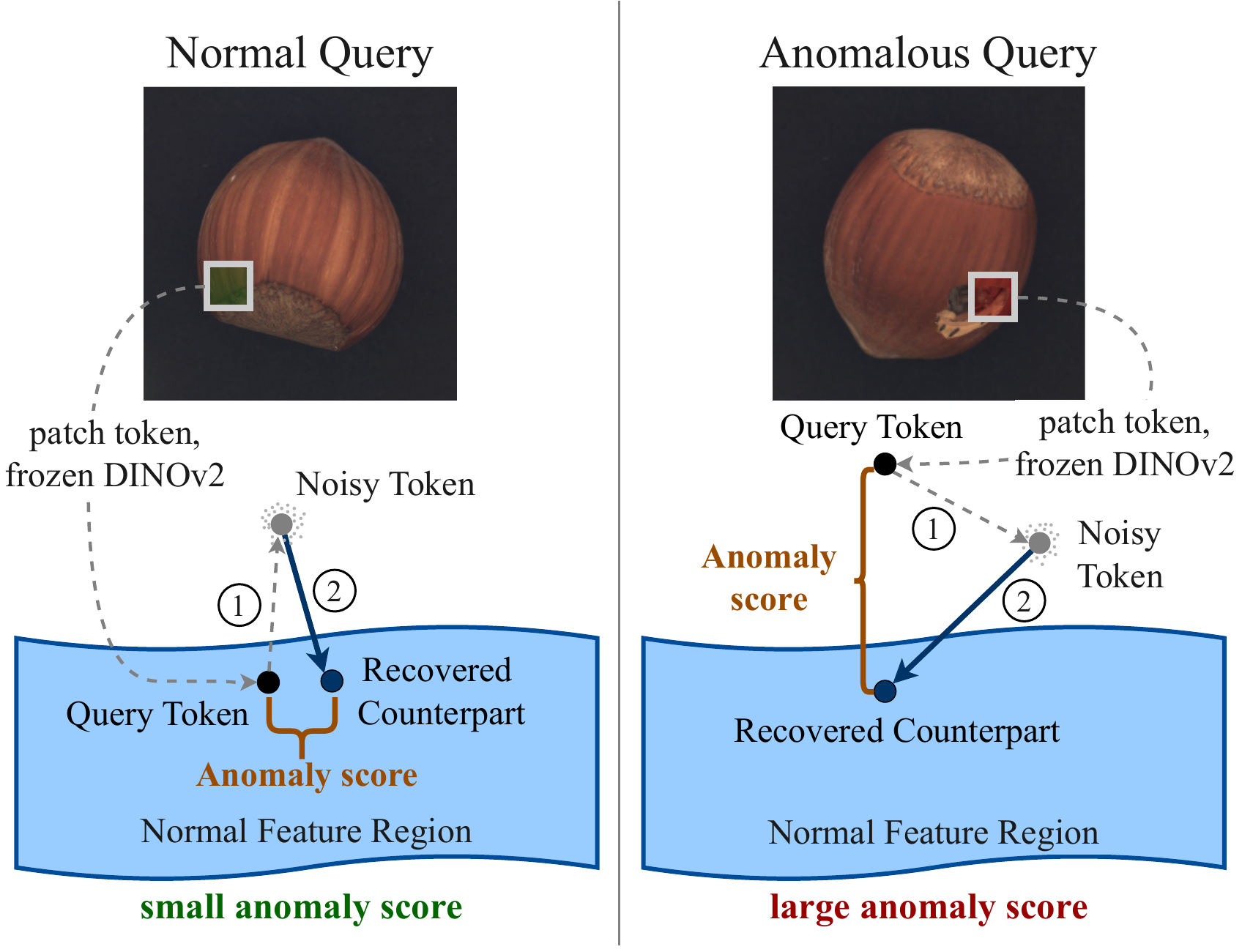}
\caption{
Conceptual illustration of Flownomaly. A query patch token is (1) perturbed with Gaussian noise and (2) recovered using a single evaluation of the velocity field fitted to the normal supports. Tokens consistent with normal appearance undergo little displacement (left),
whereas less compatible tokens tend to undergo larger changes toward the normal
support features (right). The squared distance between a query token and its recovered counterpart defines the anomaly score.
}
\label{fig:teaser}
\end{figure}

Frozen visual foundation models provide strong transferable representations in this low-data regime, but existing paradigms face distinct limitations: stored feature banks tie inference to finite exemplars~\cite{patchcore,anomalydino}, fitted linear subspaces impose rigid linearity assumptions~\cite{subspacead}, reference-graph alignment requires query-time optimization~\cite{anoco}, and feature reconstruction risks reproducing anomalies~\cite{you2022unified,fang2023fastrecon}.

To overcome these limitations, we formulate FSAD as normal-counterpart recovery in frozen feature space by asking: can few-shot normal supports define a \textit{nonlinear recovery operator} that maps a query token to a normality-biased counterpart in a single step? The discrepancy between the observed token and this recovered counterpart provides the anomaly signal. Conditional flow matching (CFM)~\cite{lipman2023flow,liu2023flow,albergo2025stochastic} provides the training principle for this operator by fitting a time-conditioned velocity field that transports Gaussian noise to normal support features. Under squared loss, the CFM optimum predicts the conditional mean path
velocity; along a straight path, this determines the posterior mean of the clean normal feature in one field evaluation, without trajectory integration.

\textbf{Flownomaly} applies this recovery principle to standardized multi-depth tokens from a frozen DINOv2 encoder (Fig.~\ref{fig:teaser}). Training the velocity field exclusively on normal tokens biases the recovery toward normal appearance rather than reproducing anomalous content. At inference, all query tokens are perturbed at a calibrated operating time and recovered in a single step via one batched velocity-field evaluation. However, single-step recovery introduces a calibration challenge: the operating time balances sensitivity to query variations against
upper-tail residual variation on normal features. In the absence of anomalous validation samples in FSAD, the operating time should be chosen from normal supports alone. Flownomaly addresses this via a normal-only \textit{Jacobian gain-to-noise} criterion. Support-derived augmentations also drive field fitting and checkpoint selection, ensuring that no stage relies on anomalous samples, auxiliary supervision, or test labels.

Experiments on MVTec-AD~\cite{mvtec}, VisA~\cite{visa}, and Real-IAD~\cite{realiad} demonstrate strong detection and localization performance, with the largest image-level gains on VisA and Real-IAD. Ablations further show that trajectory integration provides no detection advantage, reveal a broad region of well-performing operating times, and show that the normal-only criterion selects within it. In summary, this paper provides the following contributions:
\begin{itemize}
\item A formulation of FSAD as \textit{normal-counterpart recovery}
in frozen feature space, using a single-step recovery operator derived from
the posterior-mean identity of straight-path CFM.
\item A normal-only \textit{Jacobian gain-to-noise} calibration criterion that
selects the operating time from normal supports alone, without anomalous
validation data or test labels.
\item An extensive evaluation across three benchmarks demonstrating state-of-the-art results and showing that single-step recovery matches trajectory integration at substantially lower velocity-network cost.
\end{itemize}
\section{Related Work}
\label{sec:related_work}

\subsection{Few-Shot Anomaly Detection}

Existing FSAD frameworks differ primarily in how normal supports are modeled and compared against query features. Reference-based approaches directly retain normal features for inference: PatchCore~\cite{patchcore} constructs a coreset-reduced patch memory, AnomalyDINO~\cite{anomalydino} performs nearest-neighbor matching over frozen DINOv2 tokens, and WinCLIP~\cite{winclip} combines visual-reference association with vision-language alignment. Beyond direct feature matching, PromptAD~\cite{promptad} learns category-specific one-class prompts from few-shot normal supports, while registration-based methods~\cite{huang2022registration} learn category-agnostic correspondences between query and reference features that transfer to unseen object categories.

More structured frameworks explicitly model normal feature geometry and transformations. SubspaceAD~\cite{subspacead} represents normal DINOv2 tokens through a linear principal-component subspace, DCP-SFR~\cite{dcpsfr} adds structure-aware refinement to preserve weak anomaly cues, ANoCo~\cite{anoco} measures the optimization displacement required to align queries with an anchored reference graph, and FoundAD~\cite{zhai2026foundation} learns a nonlinear projection onto the natural-image manifold in a unified multi-class setting.

In contrast to discrete memory banks or rigid linear subspaces, Flownomaly represents target-category normality through a continuous, time-conditioned velocity field over frozen features, eliminating nearest-neighbor search and query-time optimization.

\subsection{Normal-Counterpart Recovery}

Reconstruction-based models score discrepancies between observed query features and their normal counterparts. A central challenge lies in preventing the network from reproducing anomalous content. Reverse Distillation~\cite{deng2022anomaly} reconstructs multi-scale teacher features through a one-class bottleneck, while UniAD~\cite{you2022unified} employs neighbor-masked attention and feature jittering to limit this identity shortcut. In FSAD, FastRecon~\cite{fang2023fastrecon} reconstructs query features from normal references through distribution-regularized regression, FastRef~\cite{li2026fastref} transfers query characteristics into normal prototypes while suppressing anomalous content, and INP-Former~\cite{inpformer} extracts normal prototypes from the query image as linear combinations of its own tokens, scoring residuals through a prototype-guided decoder.

Generative models obtain normal counterparts through denoising. DiffusionAD~\cite{zhang2025diffusionad} performs one-step noise-to-normal image reconstruction scored by a segmentation network, while OmiAD~\cite{feng2025omiad} distills an adaptive masked diffusion model into a one-step multi-class detector.

Flownomaly instead learns feature-space recovery from the support set alone, keeping the normality model independent of query content. The method requires no synthetic abnormal--normal pairs, teacher distillation, pixel decoders, or auxiliary segmentation heads.

\subsection{Flow-Based Anomaly Detection}

Flow-based anomaly detection traditionally models normal data through
invertible mappings to tractable latent priors. Normalizing flows derive
anomaly scores from exact likelihood estimates or latent log-probabilities, as
in CFLOW-AD~\cite{gudovskiy2022cflow}, FastFlow~\cite{yu2021fastflow}, and
MSFlow~\cite{zhou2024msflow}.

Recent flow-matching methods instead use learned velocity fields for
correction or anomaly scoring. In brain imaging,
Reflect~\cite{BeiFar_Reflect_MICCAI2025} learns a one-step rectified-flow
correction from synthetically corrupted to clean VAE latents.
WT-Flow~\cite{li2025time} combines time-reversed flow matching with a local
Worst Transport interpolant to estimate a density proxy.
TCCM~\cite{li2026scalable} scores deviations from time-conditioned contraction
dynamics, whereas Flow Mismatching~\cite{chen2026flow} aggregates velocity
discrepancies across multiple query paths.

Unlike prior flow-based detectors that rely on likelihood proxies, reverse
trajectories, or multi-path velocity discrepancies, Flownomaly employs flow
matching exclusively for normal-counterpart recovery, eliminating trajectory
integration and multi-path aggregation. In practice, one batched velocity-field
evaluation per image produces the residuals used for both dense localization
and image-level detection.

\section{Method}
\label{sec:method}

\begin{figure*}[t]
\centering
\includegraphics[width=\linewidth]{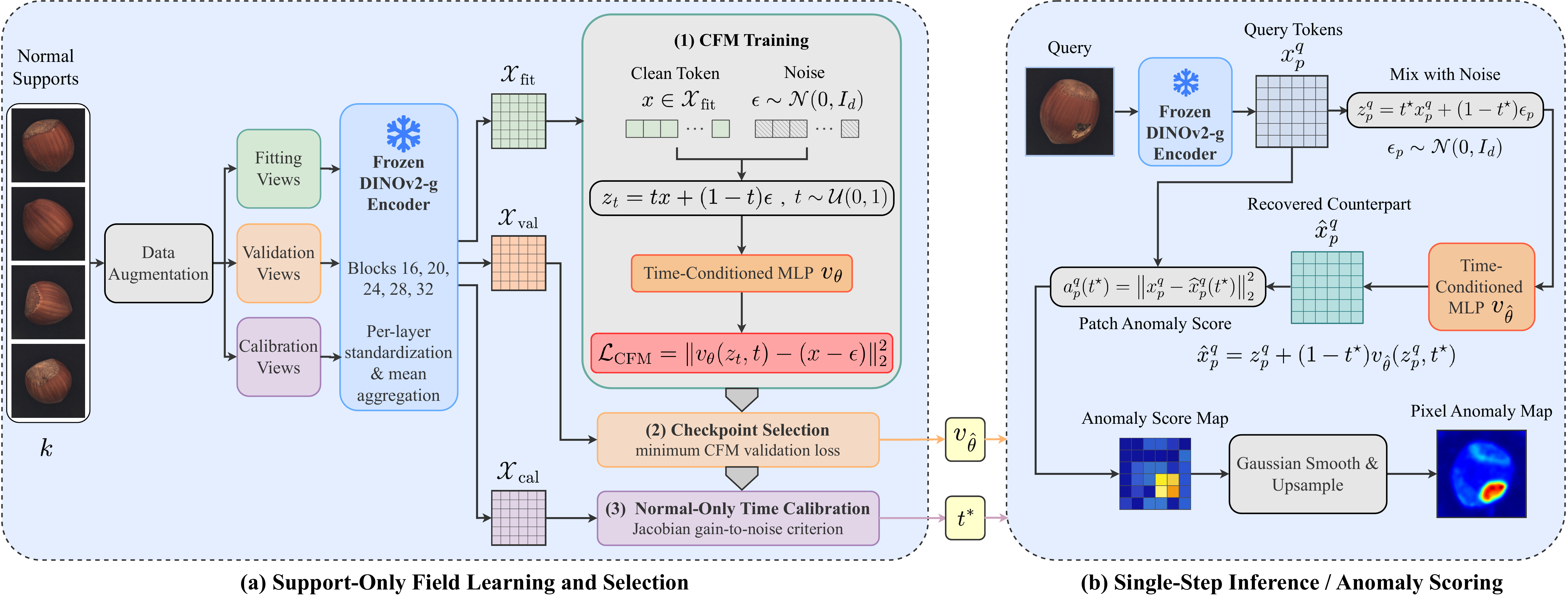}
\caption{
Overview of Flownomaly.
(a)~Disjoint augmented views of the \(k\) normal supports are encoded into
standardized multi-depth DINOv2 tokens (Sec.~\ref{sec:problem_setup}): fitting views train the velocity field
\(v_\theta\) via conditional flow matching (Sec.~\ref{sec:cfm}), validation
views select the checkpoint \(v_{\widehat\theta}\), and calibration views
determine the operating time \(t^\star\) via a normal-only criterion
(Sec.~\ref{sec:normal_only_selection}).
(b)~At inference, query tokens perturbed with Gaussian noise at \(t^\star\)
are recovered in one batched evaluation of \(v_{\widehat\theta}\) (Sec.~\ref{sec:one_step_recovery}). Squared
residuals to recovered counterparts form the anomaly score map, which is
smoothed and upsampled for dense localization (Sec.~\ref{sec:scoring}).
}
\label{fig:method_overview}
\end{figure*}

Flownomaly is a category-specific few-shot anomaly detection method that models normality
through a nonlinear recovery operator over frozen DINOv2 patch tokens. Figure~\ref{fig:method_overview} provides an overview of the method. Given \(k\)
defect-free support images alongside an unlabelled query image, Flownomaly predicts
an image-level anomaly score and a dense pixel-level localization map. Support views are
encoded into standardized multi-depth features, and a compact time-conditioned velocity
MLP is trained via conditional flow matching (CFM) along straight paths to transport
Gaussian noise toward normal tokens. Under squared-error CFM, the population-optimal field predicts the conditional mean path velocity, and along a straight path its single-step update yields a
posterior-mean normal counterpart without trajectory integration.

At inference, query tokens perturbed at a calibrated operating time \(t^\star\) are recovered in one batched forward pass. Tokens consistent with normal supports typically undergo little displacement, whereas anomalous tokens tend to receive larger normal-directed updates, yielding residuals that delineate defects. Checkpoint selection and operating-time calibration rely exclusively on support-derived normal data.
The remainder of this section details support feature extraction (Sec.~\ref{sec:problem_setup}),
velocity field fitting (Sec.~\ref{sec:cfm}), single-step recovery
(Sec.~\ref{sec:one_step_recovery}), anomaly scoring (Sec.~\ref{sec:scoring}), and
normal-only operating-time calibration (Sec.~\ref{sec:normal_only_selection}).

\subsection{Problem Setup and Support Features}
\label{sec:problem_setup}

Each category provides a support set of \(k\in\{1,2,4\}\) defect-free images. Given an unlabelled query image \(I^q\), the objective is to predict both an image anomaly score and a dense pixel-level localization map.

\smallskip
\noindent\textbf{Multi-depth DINOv2 features.}
Following SubspaceAD~\cite{subspacead}, Flownomaly represents each patch with tokens from middle-to-late blocks of a frozen DINOv2 encoder~\cite{dinov2} rather than the final layer alone, capturing complementary texture and semantic cues. The multi-depth token \(x_p(I)\in\mathbb R^d\) at position \(p\), where \(d\) is
the encoder's token dimension, is formed by standardizing each block's features with channel-wise statistics estimated on fitting views alone, then averaging across the selected blocks (Suppl.~\ref{supp:support_sampling}). Standardizing prior to averaging aligns feature scales across depths and prevents blocks with larger activations from dominating.

\smallskip
\noindent\textbf{Support-derived normal pools.}
The few-shot normal supports should fulfill three distinct roles: fitting the
velocity field, guiding checkpoint selection, and calibrating the operating time.
Holding out entire images is impossible in the one-shot (\(k=1\)) setting;
instead, three disjoint sets of augmentations of the same \(k\) support images
are generated via separate random streams. Valid tokens from these streams form the
fitting, validation, and calibration pools \(\mathcal X_{\mathrm{fit}}\),
\(\mathcal X_{\mathrm{val}}\), and \(\mathcal X_{\mathrm{cal}}\), providing
held-out augmented views for checkpoint selection and calibration. Augmentation padding is discarded; view counts and random stream details are discussed in Sec.~\ref{sec:implementation} and Suppl.~\ref{supp:support_sampling}.

\subsection{Conditional Flow Matching on Normal Tokens}
\label{sec:cfm}

To represent the normality of the target category, Flownomaly learns a
time-conditioned velocity field that transports Gaussian noise toward the
distribution of normal support features; once trained, this field enables single-step
counterpart recovery (Sec.~\ref{sec:one_step_recovery}). Fitting follows
conditional flow matching along the straight path between noise and
features~\cite{lipman2023flow,liu2023flow,albergo2025stochastic}. For a fitting token
\(x\in\mathcal X_{\mathrm{fit}}\), noise \(\epsilon\sim\mathcal N(0,I_d)\), and a
time \(t\sim\mathcal U(0,1)\), the training state is
\begin{equation}
z_t=tx+(1-t)\epsilon .
\label{eq:cfm_path}
\end{equation}
The state \(z_t\in\mathbb R^d\) interpolates between Gaussian noise at \(t=0\) and a
clean normal token at \(t=1\) along a straight trajectory with constant velocity
\(x-\epsilon\). Sampling \(t\sim\mathcal U(0,1)\) uniformly trains the field across all
noise-to-feature perturbation scales.
Given a state \(z\) and time \(t\), the velocity field
\(v_\theta(z,t)\in\mathbb R^d\), parameterized by weights \(\theta\), predicts
the velocity of the noise-to-feature interpolation.
Implemented as a residual MLP with sinusoidal time embeddings (detailed in Suppl.~\ref{supp:architecture_optimization}), it
represents the sole trained component, as the encoder remains frozen. Its
parameters minimize the training objective:
\begin{equation}
\mathcal L_{\mathrm{CFM}}(\theta)=
\mathbb E_{x,\epsilon,t}
\left[
\left\|v_\theta(z_t,t)-(x-\epsilon)\right\|_2^2
\right].
\label{eq:cfm_objective}
\end{equation}

\smallskip
\noindent\textbf{Support-only checkpoint selection.}
During training, each checkpoint is evaluated on
\(\mathcal X_{\mathrm{val}}\) using fixed noise and interpolation-time grids.
The checkpoint with the lowest validation CFM loss is selected, and its
parameters are denoted by \(\widehat{\theta}\). The selected field is then used for
operating-time calibration and inference.

\subsection{Single-Step Conditional-Mean Recovery}
\label{sec:one_step_recovery}
CFM learns a velocity field along the full noise-to-normal path, whereas
Flownomaly requires only a normality-biased counterpart for each query token.
For the straight interpolation in Eq.~\eqref{eq:cfm_path}, the posterior mean
of the clean normal token is available from a single evaluation of the
population-optimal field at an intermediate state.

Let \(X\) be a normal token from the empirical fitting distribution,
let \(\epsilon\sim\mathcal N(0,I_d)\) be independent Gaussian noise, and let \(Z_t=tX+(1-t)\epsilon\) denote the corresponding random
interpolation state.
Under the squared-error objective in Eq.~\eqref{eq:cfm_objective}, the
population-optimal field is the conditional expectation of its regression
target~\cite{lipman2023flow,albergo2025stochastic}:
\begin{equation}
v^\star(z,t)
=
\mathbb E\!\left[X-\epsilon\mid Z_t=z\right].
\label{eq:optimal_velocity}
\end{equation}
From the definition of \(Z_t\),
\begin{equation}
X=Z_t+(1-t)(X-\epsilon).
\label{eq:state_rearrangement}
\end{equation}
Taking conditional expectations given \(Z_t=z\), using
\(\mathbb E[Z_t\mid Z_t=z]=z\) and applying
Eq.~\eqref{eq:optimal_velocity}, results in
\begin{equation}
\mathbb E\!\left[X\mid Z_t=z\right]
=
z+(1-t)v^\star(z,t).
\label{eq:posterior_mean}
\end{equation}
At the population optimum, this single-step recovery is a conditional
clean-token estimator expressed through the straight-path CFM velocity field.
This identity relates to Tweedie's formula~\cite{efron2011tweedie} and to
conversions between velocity, noise, and clean-data prediction in diffusion
models~\cite{ho2020denoising,salimans2022progressive}.
Flownomaly uses this relation to construct a nonlinear, normality-biased
recovery operator from few-shot normal supports. The learned field is evaluated
once at a perturbed query state, and the discrepancy between the query token
and its recovered counterpart provides the anomaly score.
Replacing \(v^\star\) with the fitted field \(v_{\widehat{\theta}}\) defines the recovery operator
\begin{equation}
\mathcal R_{\widehat{\theta},t}(z)
=
z+(1-t)v_{\widehat{\theta}}(z,t).
\label{eq:recovery_operator}
\end{equation}
The update spans the remaining interval \(1-t\) to the data endpoint \(t=1\)
in one operation, termed \emph{single-step recovery}. Although equivalent to
one Euler step of size \(1-t\), it estimates the conditional mean rather than
an integrated ODE endpoint. Multi-step integration provides no detection
advantage (Sec.~\ref{sec:one_step_ablation}).

For a query image \(I^q\), let \(x_p^q=x_p(I^q)\) denote the standardized
token at position \(p\). Each query token is perturbed with
\(\epsilon_p\sim\mathcal N(0,I_d)\) at operating time \(t\),
\begin{equation}
z_p^q(t)=t\,x_p^q+(1-t)\epsilon_p,
\label{eq:perturbed_query}
\end{equation}
and its counterpart is recovered as
\begin{equation}
\widehat{x}_p^q(t)
=
\mathcal R_{\widehat{\theta},t}\!\left(z_p^q(t)\right).
\label{eq:recovered_counterpart}
\end{equation}
All query tokens are processed in one batch, so recovery requires a single
forward pass through \(v_{\widehat{\theta}}\).
Equation~\eqref{eq:posterior_mean} identifies the conditional mean under the
normal-token corruption model. For potentially anomalous queries,
Eq.~\eqref{eq:recovered_counterpart} uses the learned field to estimate a
normality-biased counterpart; the identity alone does not guarantee recovery
of an actual defect-free counterpart or anomaly separation.
The post-hoc analysis in Suppl.~\ref{supp:recovery_mechanism} supports this interpretation: recovery
residuals increase with distance to the unrotated support-token bank, while
recovered tokens generally move closer to that bank.

\subsection{Anomaly Scoring}
\label{sec:scoring}
The anomaly score for each patch token is given by the squared recovery residual:
\begin{equation}
a_p^q(t)
=
\left\|x_p^q-\widehat{x}_p^q(t)\right\|_2^2 .
\label{eq:token_score}
\end{equation}

Collecting \(a_p^q(t)\) over all patch positions forms the residual map at token resolution. For
dense localization, the map is Gaussian-smoothed with \(\sigma=0.8\) token widths
and bilinearly upsampled to image resolution to yield the predicted anomaly map. For image-level detection, the
anomaly score averages the largest \(\alpha\) fraction of the unsmoothed,
upsampled map; both predictions are evaluated at \(t^\star\).

\subsection{Normal-Only Operating-Time Calibration}
\label{sec:normal_only_selection}

The operating time \(t\) controls how strongly recovery modifies a query
token. At the population optimum, recovery becomes query-independent as
\(t\to0\) and approaches the global normal mean, while as \(t\to1\), the factor
\(1-t\) suppresses the field update and recovery approaches the identity. Neither extreme
provides useful conditional recovery, so the operating time must be selected
from the interior. Without a normal-only selection rule, \(t\) must be fixed
in advance.

Let \(\rho_t(x;\epsilon)=x-\mathcal R_{\widehat{\theta},t}
(tx+(1-t)\epsilon)\) denote the recovery residual, whose squared norm is the
anomaly score of Eq.~\eqref{eq:token_score}. A useful operating time makes this
residual respond strongly to small deviations from normal appearance while
remaining small on normal tokens.

Locally, an anomalous token can be viewed as a perturbed normal token. This
motivates a signal-to-noise criterion computed from normal calibration views
alone: the residual response to a perturbation is the signal, while the
upper-tail residual of normal tokens is the noise level.

For calibration view \(m\) and noise repetition \(j\), the gain
\(\mathcal G_{m,j}(t)\) measures the squared first-order response of the
residual to input changes, averaged over calibration tokens and
random unit directions, since anomaly directions are unknown. The upper-tail
level \(\mathcal F_{m,j}(t)\) is the mean of the largest \(1\%\) of squared
residuals produced by the same tokens. Their ratio defines the Jacobian
gain-to-noise (JGN) score:
\begin{equation}
\mathcal C_{m,j}(t)
=
\frac{\mathcal G_{m,j}(t)}
     {\mathcal F_{m,j}(t)+\delta},
\label{eq:jvp_gain_noise}
\end{equation}
where \(\delta>0\) provides numerical stability. Under a first-order isotropic
approximation, a larger JGN score indicates that smaller input deviations can
produce increases in squared residuals comparable to the upper-tail squared
residuals of normal tokens.

Over the candidate time grid \(\mathcal T\), the selected operating time
\(t^\star\) is the smallest maximizer of the median across calibration views
of the per-view median across noise repetitions. Median aggregation limits
the influence of atypical views and noise draws. The same random draws are reused across candidate times, and no
anomalous samples or test labels are used. JGN is therefore a normal-only rule
for selecting the operating time rather than an estimator of the category-wise
test-label optimum. Formal definitions, the random-probe estimator, and the
selection rule are given in Suppl.~\ref{supp:calibration_scoring}; the candidate grid and remaining
constants appear in Sec.~\ref{sec:implementation}.
\section{Experiments}
\label{sec:experiments}

The experiments evaluate Flownomaly for few-shot anomaly detection and localization on MVTec-AD and VisA (Sec.~\ref{sec:main_results}), and under the sample-shot setting on Real-IAD (Sec.~\ref{sec:real_iad}). Beyond baseline comparisons, targeted ablations validate Flownomaly's central design choices, spanning single-step recovery, normal-only operating-time calibration, residual scoring, and backbone scaling (Sec.~\ref{sec:ablation_study}).
\begin{table*}[!t]
\centering
\caption{
    Few-shot anomaly detection performance on MVTec-AD and VisA for category-specific methods. Flownomaly is reported as mean \(\pm\) standard deviation over 3 seeds.
    All values are percentages.
    \textbf{Bold} and \underline{underlined} indicate the best and second-best results, respectively.
}
\label{tab:few_shot_mvtec_visa}

\vspace{2pt}

\begingroup
\renewcommand{\resultpm}[2]{%
    \ensuremath{#1_{\scriptstyle \pm #2}}%
}
\renewcommand{\bestpm}[2]{%
    \ensuremath{\bm{#1}_{\scriptstyle \pm #2}}%
}
\renewcommand{\secondbestpm}[2]{%
    \underline{\ensuremath{#1_{\scriptstyle \pm #2}}}%
}
\resizebox{\textwidth}{!}{%
\renewcommand{\arraystretch}{1.08}%
\setlength{\tabcolsep}{5pt}%
\begin{tabular}{c l|ccc|ccc|ccc|ccc}
\toprule

\multirow{3}{*}{\textbf{Setup}}
& \multirow{3}{*}{\textbf{Method}}
& \multicolumn{6}{c|}{\textbf{MVTec-AD}}
& \multicolumn{6}{c}{\textbf{VisA}}
\\
\cmidrule(lr){3-8}\cmidrule(lr){9-14}

&
& \multicolumn{3}{c|}{\textbf{Image-level}}
& \multicolumn{3}{c|}{\textbf{Pixel-level}}
& \multicolumn{3}{c|}{\textbf{Image-level}}
& \multicolumn{3}{c}{\textbf{Pixel-level}}
\\

&
& AUROC & AUPR & \(F_{1,\max}\)
& AUROC & AUPRO & \(F_{1,\max}\)
& AUROC & AUPR & \(F_{1,\max}\)
& AUROC & AUPRO & \(F_{1,\max}\)
\\

\midrule

% =========================================================
% 1-shot
% =========================================================

\multirow{8}{*}{\textbf{1-shot}}
& PatchCore~\cite{patchcore}
& 83.4 & 92.2 & 90.5
& 92.0 & 79.7 & 50.4
& 79.9 & 82.8 & 81.7
& 95.4 & 80.5 & 38.0
\\

& WinCLIP~\cite{winclip}
& 93.1 & 96.5 & 93.7
& 95.2 & 87.1 & 55.9
& 83.8 & 85.1 & 83.1
& 96.4 & 85.1 & 41.3
\\

& PromptAD~\cite{promptad}
& 94.6 & 97.1 & 95.2
& 95.9 & 87.9 & 53.2
& 86.9 & 88.4 & 82.9
& 96.7 & 85.1 & 35.0
\\

& AnomalyDINO~\cite{anomalydino}
& 96.6 & 98.2 & 95.8
& 96.8 & 92.7 & 60.2
& 87.4 & 89.0 & 84.3
& 97.8 & 92.5 & 45.1
\\

& SubspaceAD~\cite{subspacead}
& 97.1 & 98.6 & 96.7
& 97.5 & 94.5 & 60.5
& \secondbest{93.2} & 93.0 & \secondbest{89.6}
& 98.2 & \secondbest{95.5} & 47.5
\\

& DCP-SFR~\cite{dcpsfr}
& 97.3 & 98.8 & 96.7
& 96.9 & 92.7 & 60.7
& 92.0 & \secondbest{93.7} & 87.8
& 98.2 & 93.3 & 44.9
\\

& ANoCo~\cite{anoco}
& \best{97.9} & \best{99.1} & \secondbest{96.9}
& \secondbest{97.7} & \secondbest{95.4} & \secondbest{64.9}
& 92.7 & 93.3 & 88.8
& \best{98.7} & 94.9 & \secondbest{51.9}
\\

\cmidrule(l){2-14}
& \textbf{Flownomaly}
& \secondbestpm{97.8}{0.9} & \bestpm{99.1}{0.3} & \bestpm{97.4}{0.3}
& \bestpm{98.0}{0.1} & \bestpm{95.5}{0.3} & \bestpm{65.3}{0.5}
& \bestpm{95.3}{0.7} & \bestpm{95.6}{0.7} & \bestpm{91.7}{0.6}
& \bestpm{98.7}{0.0} & \bestpm{96.3}{0.2} & \bestpm{54.0}{0.2}
\\

\midrule

% =========================================================
% 2-shot
% =========================================================

\multirow{8}{*}{\textbf{2-shot}}
& PatchCore~\cite{patchcore}
& 86.3 & 93.8 & 92.0
& 93.3 & 82.3 & 53.0
& 81.6 & 84.8 & 82.5
& 96.1 & 82.6 & 41.0
\\

& WinCLIP~\cite{winclip}
& 94.4 & 97.0 & 94.4
& 96.0 & 88.4 & 58.4
& 84.6 & 85.8 & 83.0
& 96.8 & 86.2 & 43.5
\\

& PromptAD~\cite{promptad}
& 95.7 & 97.9 & 95.3
& 96.2 & 88.5 & 54.3
& 88.3 & 90.0 & 82.4
& 97.1 & 85.8 & 37.1
\\

& AnomalyDINO~\cite{anomalydino}
& 96.9 & 98.2 & 96.1
& 97.0 & 93.1 & 61.0
& 89.7 & 90.7 & 86.3
& 98.0 & 93.4 & 47.6
\\

& SubspaceAD~\cite{subspacead}
& 97.5 & 98.8 & 97.0
& 97.8 & 94.9 & 61.7
& \secondbest{93.8} & 93.4 & 90.2
& 98.3 & \secondbest{95.7} & 45.8
\\

& DCP-SFR~\cite{dcpsfr}
& 97.7 & 98.9 & 97.0
& 97.1 & 93.3 & 61.6
& 92.5 & \secondbest{94.2} & 88.0
& 98.4 & 93.8 & 46.9
\\

& ANoCo~\cite{anoco}
& \best{98.4} & \secondbest{99.2} & \secondbest{97.3}
& \secondbest{98.0} & \best{96.0} & \secondbest{65.3}
& 93.3 & 92.9 & \secondbest{90.3}
& \secondbest{98.7} & 94.7 & \secondbest{53.0}
\\

\cmidrule(l){2-14}
& \textbf{Flownomaly}
& \bestpm{98.4}{0.5} & \bestpm{99.4}{0.2} & \bestpm{97.7}{0.5}
& \bestpm{98.3}{0.0} & \bestpm{96.0}{0.2} & \bestpm{66.2}{0.2}
& \bestpm{96.1}{0.3} & \bestpm{95.9}{0.3} & \bestpm{92.8}{0.5}
& \bestpm{98.9}{0.0} & \bestpm{96.6}{0.1} & \bestpm{54.3}{0.5}
\\

\midrule

% =========================================================
% 4-shot
% =========================================================

\multirow{8}{*}{\textbf{4-shot}}
& PatchCore~\cite{patchcore}
& 88.8 & 94.5 & 92.6
& 94.3 & 84.3 & 55.0
& 85.3 & 87.5 & 84.3
& 96.8 & 84.9 & 43.9
\\

& WinCLIP~\cite{winclip}
& 95.2 & 97.3 & 94.7
& 96.2 & 89.0 & 59.5
& 87.3 & 88.8 & 84.2
& 97.2 & 87.6 & 47.0
\\

& PromptAD~\cite{promptad}
& 96.6 & 98.5 & 96.0
& 96.5 & 90.5 & 55.2
& 89.1 & 90.8 & 83.1
& 97.4 & 86.2 & 38.9
\\

& AnomalyDINO~\cite{anomalydino}
& 97.7 & 98.7 & 96.6
& 97.2 & 93.4 & 61.8
& 92.6 & 92.9 & 88.8
& 98.2 & 94.1 & 49.4
\\

& SubspaceAD~\cite{subspacead}
& 98.0 & 99.0 & 97.2
& 97.9 & 95.1 & 62.7
& 94.7 & 94.3 & 91.1
& 98.4 & \secondbest{96.0} & 47.1
\\

& DCP-SFR~\cite{dcpsfr}
& 98.0 & 98.9 & 96.8
& 97.5 & 93.5 & 61.9
& 92.8 & 94.7 & 88.9
& 98.6 & 94.2 & 47.1
\\

& ANoCo~\cite{anoco}
& \best{98.7} & \secondbest{99.3} & \secondbest{97.6}
& \secondbest{98.1} & \best{96.2} & \secondbest{65.7}
& \secondbest{95.2} & \secondbest{95.1} & \secondbest{91.7}
& \secondbest{98.8} & 95.7 & \best{54.2}
\\

\cmidrule(l){2-14}
& \textbf{Flownomaly}
& \secondbestpm{98.5}{0.4} & \bestpm{99.4}{0.2} & \bestpm{97.9}{0.5}
& \bestpm{98.4}{0.1} & \bestpm{96.2}{0.2} & \bestpm{66.6}{0.4}
& \bestpm{96.3}{0.4} & \bestpm{95.9}{0.6} & \bestpm{93.1}{0.5}
& \bestpm{99.0}{0.0} & \bestpm{96.7}{0.0} & \bestpm{54.2}{0.3}
\\

\bottomrule
\end{tabular}%
}
\endgroup
\end{table*}

\subsection{Datasets}
\label{sec:datasets}
Flownomaly is evaluated on three industrial anomaly detection benchmarks: MVTec-AD~\cite{mvtec}, VisA~\cite{visa}, and Real-IAD~\cite{realiad}, comprising 15, 12, and 30 categories, respectively. 
Following established FSAD protocols~\cite{anomalydino,subspacead,anoco}, the
first two benchmarks use randomly selected normal reference images under 1-, 2-,
and 4-shot settings. For Real-IAD, each selected physical sample contributes five fixed views under
a multi-view sample-shot protocol.

\subsection{Evaluation Metrics}
\label{sec:metrics}
Following standard practice~\cite{patchcore,anomalydino}, image-level performance is evaluated by AUROC (I-AUROC), AUPR, and \(F_{1,\max}\), while pixel-level performance is evaluated via AUROC (P-AUROC),
AUPRO, and \(F_{1,\max}\), with all metrics averaged across categories. \(F_{1,\max}\) is maximized over decision thresholds, and AUPRO is integrated up to a false-positive rate of \(0.3\)~\cite{mvtec}.

\subsection{Implementation Details}
\label{sec:implementation}
Flownomaly employs a frozen DINOv2 ViT-g/14-reg encoder~\cite{dinov2,darcet2024vision} at \(672\times672\) resolution, aggregating features from blocks \(\{16,20,24,28,32\}\), corresponding to 40--80\% relative
depth. The encoder output is a
\(48\times48\) spatial grid of standardized patch tokens. The time-conditioned
residual MLP uses a bottleneck width of \(d/2\), half the token embedding
dimension, and is optimized with AdamW and a cosine learning-rate
schedule. Following SubspaceAD~\cite{subspacead}, each normal support is augmented with 30 random rotations sampled from \(0^\circ\)--\(345^\circ\) for fitting, while validation and calibration each use three independently sampled rotations. Consistent with prior work~
\cite{anomalydino,subspacead}, rotation augmentation is disabled for the
orientation-sensitive MVTec-AD \textit{transistor} category, which remains included in
the aggregate evaluation.

\smallskip
\noindent\textbf{Operating time and scoring.}
JGN selects \(t^\star\) from
\(\mathcal T=\{0.05,0.10,\ldots,0.95\}\) using normal calibration views
(Suppl.~\ref{supp:calibration_scoring}). Unless stated otherwise, inference uses one Gaussian sample per
token and one batched velocity-field evaluation. The image score averages the
largest \(\alpha=1\%\) of the unsmoothed, upsampled residual map, following
DINOv2-based FSAD methods~\cite{anomalydino,subspacead}.

\subsection{Baselines}
\label{sec:baselines}
Flownomaly is evaluated against representative FSAD methods spanning distinct
modeling paradigms: memory-based patch retrieval (PatchCore~\cite{patchcore},
AnomalyDINO~\cite{anomalydino}), vision-language association and prompt learning
(WinCLIP~\cite{winclip}, PromptAD~\cite{promptad}), linear subspace modeling
(SubspaceAD~\cite{subspacead}), structural cue refinement (DCP-SFR~\cite{dcpsfr}),
and graph-based alignment (ANoCo~\cite{anoco}).

\subsection{Few-Shot Anomaly Detection Performance}
\label{sec:main_results}

Table~\ref{tab:few_shot_mvtec_visa} compares Flownomaly with recent category-specific methods across the 1-, 2-, and 4-shot settings. Flownomaly consistently achieves strong localization performance across both MVTec-AD and VisA datasets, attaining the best or tied-best pixel-level results across all evaluated shot settings and metrics. The largest image-level gains occur on VisA, where AUROC surpasses the strongest baseline by \(2.1\), \(2.3\), and
\(1.1\) percentage points (pp) in the 1-, 2-, and 4-shot regimes, respectively. On
MVTec-AD, Flownomaly leads or ties seven of nine image-level comparisons and
remains within \(0.2\)~pp of ANoCo on the other two. In the full-shot setting, Flownomaly achieves the best or tied-best result in 11 of 12 metric--dataset comparisons. Complete full-shot results for MVTec-AD and VisA
are reported in Suppl.~\ref{supp:full_shot}.

\begin{figure*}[!t]
\centering
\includegraphics[width=\linewidth]{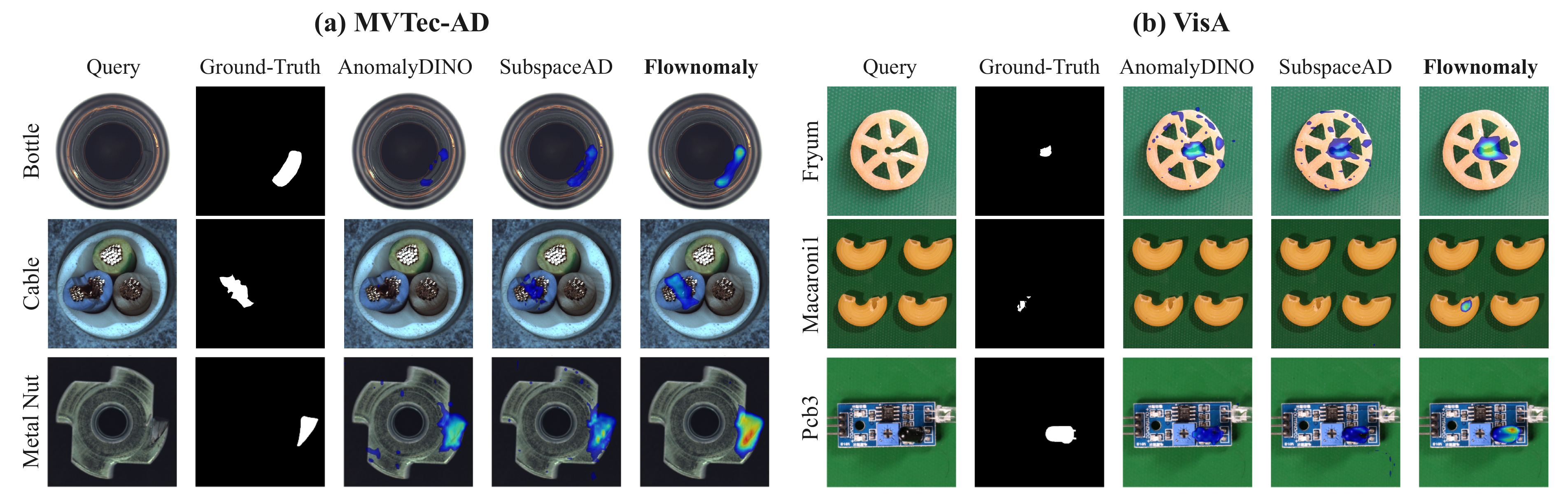}
\caption{
Qualitative comparison on MVTec-AD and VisA under the 1-shot setting. Columns
show the query image, ground-truth mask, and anomaly maps from the open-source
category-specific baselines AnomalyDINO~\cite{anomalydino} and
SubspaceAD~\cite{subspacead}, followed by Flownomaly. Flownomaly delineates
defects more sharply, with fewer spurious background activations and closer
agreement with the ground-truth masks on both datasets. Additional qualitative
results and representative failure modes are provided in Suppl.~\ref{supp:qualitative_results} and~\ref{supp:failure_modes},
respectively.
}
\label{fig:qualitative_comparison}
\end{figure*}

\smallskip
\noindent\textbf{Qualitative comparison.}
Figure~\ref{fig:qualitative_comparison} illustrates representative localization results in the 1-shot setting.  Flownomaly accurately localizes both
fine texture defects and structural damage while suppressing background
responses.

\subsection{Sample-Shot Anomaly Detection on Real-IAD}
\label{sec:real_iad}
Real-IAD contains five views per physical sample. Under the sample-shot protocol, selecting \(s\in\{1,2,4\}\) normal samples provides \(5s\) reference images. Table~\ref{tab:real_iad} reports detection and localization performance under this setting across 3 seeds. PatchCore,
PromptAD, AnomalyDINO, and
SubspaceAD are reproduced under published settings.
ANoCo and DCP-SFR are omitted as their implementations were
unavailable.

Flownomaly achieves the strongest image-level performance and the highest
pixel-level AUROC and \(F_{1,\max}\) at every sample-shot setting. In the
1-sample setting, image AUROC increases from 85.2\% with SubspaceAD to 88.2\%, while pixel-level \(F_{1,\max}\) improves from 42.7\% to 50.6\%. These results indicate that the recovery formulation remains effective under multi-view references.
 
\begin{table}[!ht]
\centering
\caption{
    Sample-shot anomaly detection and localization on Real-IAD. Results are percentages reported as mean \(\pm\) standard deviation over 3 seeds.
    \textbf{Bold} and \underline{underlined} indicate the best and second-best results, respectively.
}
\label{tab:real_iad}
\begingroup
\renewcommand{\resultpm}[2]{%
    \ensuremath{#1_{\scriptstyle \pm #2}}%
}
\renewcommand{\bestpm}[2]{%
    \ensuremath{\bm{#1}_{\scriptstyle \pm #2}}%
}
\renewcommand{\secondbestpm}[2]{%
    \underline{\ensuremath{#1_{\scriptstyle \pm #2}}}%
}
\resizebox{\columnwidth}{!}{%
\renewcommand{\arraystretch}{1.1}%
\setlength{\tabcolsep}{5.2pt}%
\begin{tabular}{l|ccc|ccc}
\toprule
& \multicolumn{3}{c|}{Image-level} & \multicolumn{3}{c}{Pixel-level} \\
Method & AUROC & AUPR & \(F_{1,\max}\) & AUROC & AUPRO & \(F_{1,\max}\) \\
\midrule
\multicolumn{7}{l}{\textbf{1-sample shot} \textit{(5 views)}} \\
\midrule
PatchCore~\cite{patchcore}     & \resultpm{79.5}{0.1} & \resultpm{77.7}{0.1} & \resultpm{71.5}{0.0} & \resultpm{96.9}{0.0} & \resultpm{84.3}{0.1} & \resultpm{30.7}{0.2} \\
PromptAD~\cite{promptad}       & \resultpm{79.4}{0.2} & \resultpm{77.2}{0.4} & \resultpm{71.4}{0.1} & \resultpm{95.9}{0.1} & \resultpm{81.7}{0.1} & \resultpm{30.8}{0.2} \\
AnomalyDINO~\cite{anomalydino} & \resultpm{82.5}{0.1} & \resultpm{78.3}{0.4} & \resultpm{73.9}{0.1} & \resultpm{97.6}{0.0} & \resultpm{91.7}{0.2} & \resultpm{37.0}{0.5} \\
SubspaceAD~\cite{subspacead}   & \secondbestpm{85.2}{0.3} & \secondbestpm{81.6}{0.4} & \secondbestpm{76.1}{0.0} & \secondbestpm{98.8}{0.0} & \bestpm{97.1}{0.1}   & \secondbestpm{42.7}{0.2} \\
\cmidrule(l){1-7}
\textbf{Flownomaly}            & \bestpm{88.2}{0.2}   & \bestpm{84.9}{0.3}   & \bestpm{78.5}{0.2}   & \bestpm{99.2}{0.0}   & \bestpm{97.1}{0.0}   & \bestpm{50.6}{0.1} \\
\midrule
\multicolumn{7}{l}{\textbf{2-sample shot} \textit{(10 views)}} \\
\midrule
PatchCore~\cite{patchcore}     & \resultpm{81.6}{0.4} & \resultpm{79.7}{0.2} & \resultpm{72.9}{0.1} & \resultpm{97.3}{0.0} & \resultpm{85.9}{0.1} & \resultpm{31.7}{0.4} \\
PromptAD~\cite{promptad}       & \resultpm{80.6}{0.5} & \resultpm{78.5}{0.7} & \resultpm{72.3}{0.4} & \resultpm{96.2}{0.1} & \resultpm{82.6}{0.4} & \resultpm{31.7}{0.7} \\
AnomalyDINO~\cite{anomalydino} & \resultpm{84.2}{0.1} & \resultpm{80.1}{0.2} & \resultpm{75.2}{0.2} & \resultpm{97.9}{0.0} & \resultpm{92.3}{0.0} & \resultpm{38.0}{0.3} \\
SubspaceAD~\cite{subspacead}   & \secondbestpm{86.3}{0.2} & \secondbestpm{82.8}{0.2} & \secondbestpm{77.1}{0.2} & \secondbestpm{98.9}{0.0} & \bestpm{97.2}{0.0}   & \secondbestpm{43.7}{0.2} \\
\cmidrule(l){1-7}
\textbf{Flownomaly}            & \bestpm{88.7}{0.1}   & \bestpm{85.0}{0.2}   & \bestpm{79.2}{0.1}   & \bestpm{99.2}{0.0}   & \secondbestpm{97.0}{0.1} & \bestpm{49.0}{0.1} \\
\midrule
\multicolumn{7}{l}{\textbf{4-sample shot} \textit{(20 views)}} \\
\midrule
PatchCore~\cite{patchcore}     & \resultpm{83.3}{0.4} & \resultpm{81.4}{0.2} & \resultpm{74.4}{0.3} & \resultpm{97.7}{0.0} & \resultpm{87.0}{0.1} & \resultpm{32.4}{0.4} \\
PromptAD~\cite{promptad}       & \resultpm{82.1}{0.3} & \resultpm{80.1}{0.4} & \resultpm{73.2}{0.1} & \resultpm{96.4}{0.1} & \resultpm{83.2}{0.3} & \resultpm{33.0}{0.6} \\
AnomalyDINO~\cite{anomalydino} & \resultpm{85.8}{0.1} & \resultpm{81.8}{0.1} & \resultpm{76.7}{0.2} & \resultpm{98.1}{0.0} & \resultpm{92.8}{0.1} & \resultpm{38.8}{0.3} \\
SubspaceAD~\cite{subspacead}   & \secondbestpm{87.1}{0.2} & \secondbestpm{83.7}{0.2} & \secondbestpm{77.8}{0.1} & \secondbestpm{98.9}{0.0} & \bestpm{97.3}{0.0}   & \secondbestpm{44.5}{0.2} \\
\cmidrule(l){1-7}
\textbf{Flownomaly}            & \bestpm{89.5}{0.2}   & \bestpm{86.0}{0.1}   & \bestpm{79.9}{0.1}   & \bestpm{99.2}{0.0}   & \secondbestpm{97.0}{0.0} & \bestpm{49.0}{0.3} \\
\bottomrule
\end{tabular}%
}
\endgroup
\end{table}

\subsection{Computational Efficiency}
\label{sec:efficiency}
Only the velocity network is optimized, as the encoder remains frozen.
In the 1-shot setting, fitting one category takes a median of \(24\)~s on MVTec-AD and \(28\)~s on
VisA, with \(2.6\)~GB of peak GPU memory. At
inference, the velocity network contributes \(2.2\)~ms of the total \(60.9\)~ms
image latency (Table~\ref{tab:nfe_ablation}), with most computation arising from
the frozen encoder. The velocity network adds 23.2\,M parameters per category
(\(\approx\)88\,MiB); unlike token memory banks, this footprint does not grow
with \(k\). Additional details are given in
Suppl.~\ref{supp:efficiency_details}.

\subsection{Ablation Study}
\label{sec:ablation_study}
The ablation study examines the main design choices of Flownomaly, focusing on single-step recovery, operating-time calibration, recovery-residual scoring, and backbone scaling. Further ablations covering the feature representation and the support-view
construction are provided in the \hyperref[supp:feature_ablations]{supplementary material}.

\paragraph{Single-step recovery is sufficient.}
\label{sec:one_step_ablation}

A central premise of Flownomaly is that the learned velocity field can directly serve as a conditional-mean recovery operator without integrating a complete flow trajectory. Table~\ref{tab:nfe_ablation} compares single-step recovery with Euler integration across 2, 5, 10, and 20 velocity evaluations
from the same perturbed query state. All variants share the checkpoint, operating time, and initial Gaussian draw, isolating the effect of trajectory integration.

\begin{table}[!ht]
\centering
\caption{Effect of the number of velocity-field evaluations (NFE) in the
1-shot setting.}
\label{tab:nfe_ablation}
\begingroup
\renewcommand{\resultpm}[2]{%
    \ensuremath{#1_{\scriptstyle \pm #2}}%
}
\resizebox{\columnwidth}{!}{%
\setlength{\tabcolsep}{5.75pt}%
\begin{tabular}{c|cc|cc|cc}
\toprule
& \multicolumn{2}{c|}{MVTec-AD} & \multicolumn{2}{c|}{VisA}
& \multicolumn{2}{c}{Latency (ms/image)} \\
NFE & I-AUROC & P-AUROC & I-AUROC & P-AUROC & Velocity & Total model \\
\midrule
\textbf{1} & \resultpm{97.8}{0.9} & \resultpm{98.0}{0.1} & \resultpm{95.3}{0.7} & \resultpm{98.7}{0.0} & 2.2 & 60.9 \\
\cmidrule(l){1-7}
2  & \resultpm{97.8}{1.0} & \resultpm{98.1}{0.1} & \resultpm{95.2}{0.7} & \resultpm{98.7}{0.0} & 3.7 & 62.4 \\
5  & \resultpm{97.8}{1.0} & \resultpm{98.1}{0.1} & \resultpm{95.1}{0.8} & \resultpm{98.7}{0.0} & 8.0 & 66.7 \\
10 & \resultpm{97.8}{1.0} & \resultpm{98.1}{0.1} & \resultpm{95.1}{0.8} & \resultpm{98.7}{0.0} & 15.1 & 73.8 \\
20 & \resultpm{97.8}{1.0} & \resultpm{98.1}{0.1} & \resultpm{95.1}{0.8} & \resultpm{98.7}{0.0} & 29.2 & 87.9 \\
\bottomrule
\end{tabular}
}
\endgroup
\end{table}

Increasing NFE from 1 to 20 changes image- and pixel-level AUROC by at most \(0.2\)~pp
while raising velocity-network latency by \(13.3\times\), from \(2.2\)
to \(29.2\) ms/image. Total model latency increases from \(60.9\) to
\(87.9\) ms/image. Therefore, full trajectory integration provides negligible detection benefit despite substantially higher computational cost, empirically supporting the single-step recovery formulation.

\paragraph{Operating-time robustness and normal-only selection.}
\label{sec:t_star_analysis}

\begin{figure}[!ht]
\centering
\includegraphics[width=\linewidth]{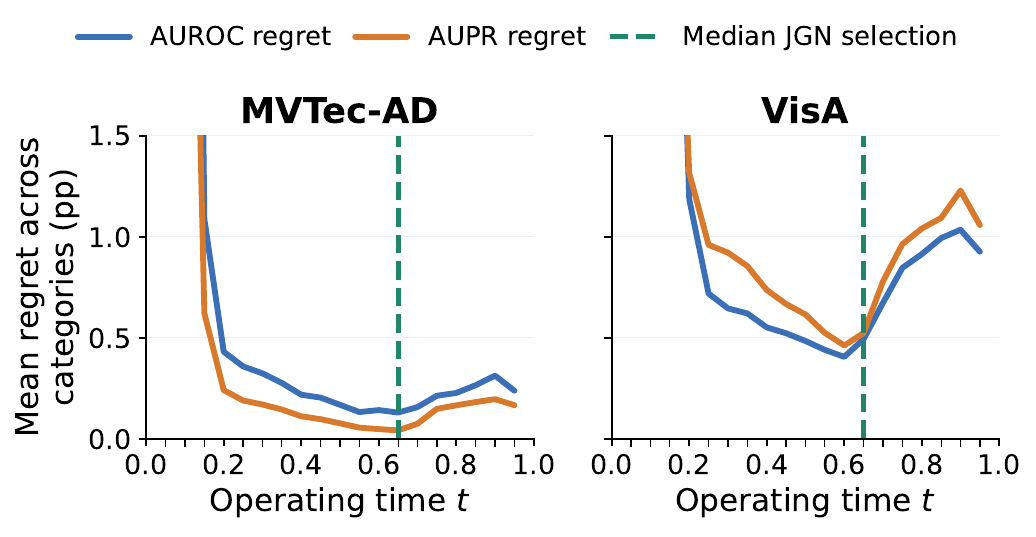}
\caption{
Operating-time robustness on MVTec-AD and VisA in the 1-shot setting. Curves
show mean category-wise image-level regret; lower is better. The dashed lines mark the median category-wise JGN selection at \(t=0.65\).
}
\label{fig:operating_time_robustness}
\end{figure}

The label-free JGN calibration is assessed by sweeping
\(t\in\mathcal T\). This reveals how sharply detection depends on \(t\) and
whether the selected time lies in a high-performing region. For each category
and metric (image-level AUROC or AUPR), the retrospective regret at \(t\) is
the gap between the best value over \(\mathcal T\) and the value at \(t\).
Figure~\ref{fig:operating_time_robustness} reports this regret averaged across
categories. Test labels are used only to define the retrospective reference.

Both datasets exhibit a broad low-regret interior, indicating robust
performance across a wide range of operating times. The median category-wise JGN selection is \(0.65\) on both datasets and lies
within this region. Category-wise JGN selection yields median image-AUROC regrets of
\(0.06\) and \(0.14\)~pp on MVTec-AD and VisA relative to the retrospective oracle. Across alternative feature configurations, JGN selects operating times from
\(0.60\) to \(0.85\) on both datasets (Suppl.~\ref{supp:feature_ablations}), rather than returning a fixed value. Additional calibration analysis is provided in Suppl.~\ref{supp:time_calibration}.

\paragraph{Recovery-residual scoring.}
\label{sec:score_semantics}

Table~\ref{tab:score_semantics} evaluates whether the proposed recovery residual
provides a more informative anomaly signal than alternative quantities derived
from the same recovery process, under otherwise identical settings. The recovery
residual explicitly compares the displacement implied by the perturbed query
with the velocity predicted by the normal model. The velocity norm measures only
the magnitude of that prediction, while the scale-invariant cosine discrepancy
measures angular disagreement between the query token and its recovered
counterpart.

\begin{table}[!ht]
\centering
\caption{Effect of the scoring rule (1-shot setting), with all other factors
fixed. Best results are in \textbf{bold}.}
\label{tab:score_semantics}
\begingroup
\renewcommand{\resultpm}[2]{%
    \ensuremath{#1_{\scriptstyle \pm #2}}%
}
\renewcommand{\bestpm}[2]{%
    \ensuremath{\bm{#1}_{\scriptstyle \pm #2}}%
}
\resizebox{\columnwidth}{!}{%
\begin{tabular}{l|cc|cc}
\toprule
& \multicolumn{2}{c|}{MVTec-AD} & \multicolumn{2}{c}{VisA} \\
Scoring rule & I-AUROC & P-AUROC & I-AUROC & P-AUROC \\
\midrule
\textbf{Recovery residual} & \bestpm{97.8}{0.9} & \bestpm{98.0}{0.1} & \bestpm{95.3}{0.7} & \bestpm{98.7}{0.0} \\
\cmidrule(l){1-5}
Velocity norm & \resultpm{96.9}{0.9} & \resultpm{97.3}{0.1} & \resultpm{94.3}{0.2} & \resultpm{98.4}{0.0} \\
Cosine residual & \resultpm{96.7}{0.6} & \resultpm{97.9}{0.1} & \resultpm{90.2}{0.4} & \resultpm{97.8}{0.2} \\
\bottomrule
\end{tabular}
}
\endgroup
\end{table}

Replacing the recovery residual with the velocity norm reduces image AUROC by
\(0.9/1.0\)~pp on MVTec-AD/VisA, while the cosine discrepancy reduces it by
\(1.1/5.1\)~pp. These results show that the full recovery discrepancy is more
informative than either alternative.

\paragraph{Backbone scaling.}
\label{sec:backbone_ablation}
The accuracy--latency trade-off is evaluated across DINOv2-S, -B, -L, and -g. For each backbone, the velocity-network bottleneck width is set to
half the encoder's token embedding dimension (\(d/2\)). Table~\ref{tab:backbone_ablation} illustrates that scaling from the DINOv2-S
configuration to the DINOv2-g improves image AUROC by \(3.1\)~pp
on MVTec-AD and \(3.5\)~pp on VisA, while model latency increases
from \(18.1\) to \(60.9\) ms/image. A comparison in Suppl.~\ref{supp:additional_backbones} shows that
DINOv3-7B~\cite{simeoni2025dinov3} does not improve over DINOv2-g in this setting.

\begin{table}[!ht]
\centering
\caption{Backbone-scale ablation in the 1-shot setting. Best results are in \textbf{bold}.}
\label{tab:backbone_ablation}
\begingroup
\renewcommand{\resultpm}[2]{%
    \ensuremath{#1_{\scriptstyle \pm #2}}%
}
\renewcommand{\bestpm}[2]{%
    \ensuremath{\bm{#1}_{\scriptstyle \pm #2}}%
}
\resizebox{\columnwidth}{!}{%
\setlength{\tabcolsep}{5.5pt}%
\begin{tabular}{lc|cc|cc}
\toprule
& & \multicolumn{2}{c|}{MVTec-AD} & \multicolumn{2}{c}{VisA} \\
Backbone & Latency (ms) & I-AUROC & P-AUROC & I-AUROC & P-AUROC \\
\midrule
DINOv2-S & 18.1 & \resultpm{94.7}{1.3} & \resultpm{97.7}{0.1} & \resultpm{91.8}{1.0} & \resultpm{98.0}{0.1} \\
DINOv2-B & 25.5 & \resultpm{96.4}{1.0} & \resultpm{97.8}{0.1} & \resultpm{93.4}{0.8} & \resultpm{98.3}{0.0} \\
DINOv2-L & 51.9 & \resultpm{97.3}{0.9} & \resultpm{97.9}{0.1} & \resultpm{94.3}{0.8} & \resultpm{98.5}{0.0} \\
\cmidrule(l){1-6}
\textbf{DINOv2-g} & 60.9 & \bestpm{97.8}{0.9} & \bestpm{98.0}{0.1} & \bestpm{95.3}{0.7} & \bestpm{98.7}{0.0} \\
\bottomrule
\end{tabular}%
}
\endgroup
\end{table}

\FloatBarrier

\raggedbottom
\section{Conclusion}
\label{sec:conclusion}
This work has presented Flownomaly, which reframes category-specific FSAD as
normal-counterpart recovery in frozen DINOv2 feature space. Its central insight
is that the conditional-mean structure of straight-path CFM enables a
normality-biased counterpart to be estimated in a single recovery step, without
trajectory integration. Residuals between query features and their recovered
counterparts provide the anomaly signal for both detection and localization.
Field fitting and checkpoint selection rely exclusively on support-derived
normal data, while the normal-only Jacobian gain-to-noise criterion calibrates
the operating time without anomalous validation data. Flownomaly achieves state-of-the-art results on VisA and Real-IAD, while leading most evaluated comparisons on MVTec-AD. Additional field evaluations provide negligible
detection benefit, while performance remains stable across a broad range of
operating times. Because token-wise residual scoring does not explicitly
enforce global structural constraints, logical anomalies such as incorrect part
counts remain challenging (Suppl.~\ref{supp:failure_modes}). Future work will investigate
category-agnostic recovery fields, explicit modeling of such global
constraints, and more compact feature extractors.
\section*{Acknowledgements}
This work is supported by the STRUCTURE ITEA 24017 project.

{
    \small
    \bibliographystyle{ieeenat_fullname}
    \bibliography{main}
}

% Append the supplement after the shared bibliography.
% Included by main.tex after the shared bibliography; not a standalone document.
\clearpage
\appendix
\setcounter{figure}{0}
\setcounter{table}{0}
\setcounter{equation}{0}
\renewcommand{\thefigure}{S\arabic{figure}}
\renewcommand{\thetable}{S\arabic{table}}
\renewcommand{\theequation}{S\arabic{equation}}
% Distinct PDF destinations are needed when counters restart in the supplement.
\renewcommand{\theHfigure}{supp.\arabic{figure}}
\renewcommand{\theHtable}{supp.\arabic{table}}
\renewcommand{\theHequation}{supp.\arabic{equation}}

\maketitlesupplementary
\phantomsection
\label{supp:material}
\addcontentsline{toc}{section}{Supplementary Material}

% Use the larger standard-deviation style consistently throughout the supplement.
\renewcommand{\resultpm}[2]{%
    \ensuremath{#1_{\scriptstyle \pm #2}}%
}
\renewcommand{\bestpm}[2]{%
    \ensuremath{\bm{#1}_{\scriptstyle \pm #2}}%
}
\renewcommand{\secondbestpm}[2]{%
    \underline{\ensuremath{#1_{\scriptstyle \pm #2}}}%
}

This supplement provides implementation, optimization, and efficiency details (Sec.~\ref{supp:implementation_and_efficiency});
additional analyses covering feature ablations, operating-time calibration, recovery contraction diagnostics, and qualitative failure modes (Sec.~\ref{supp:additional_analyses});
additional evaluations covering full-shot benchmarking and backbone comparisons (Sec.~\ref{supp:additional_evaluations});
extended qualitative localization results across datasets (Sec.~\ref{supp:qualitative_results});
and per-category few-shot and sample-shot performance tables (Sec.~\ref{supp:per_category_results}) supporting the main paper.

\section{Implementation and Efficiency}
\label{supp:implementation_and_efficiency}

\subsection{Support Sampling}
\label{supp:support_sampling}
Flownomaly results in the main paper are reported over three support samplings,
generated with random seeds 42, 43, and 44. For each sampling and category, the
selected \(k\) defect-free images determine all fitting, validation, and
calibration views.

The three pools are built from the same \(k\) images but from disjoint sets of
augmentations. The fitting pool contains each support image together with 30
rotations drawn uniformly from \(0^\circ\)--\(345^\circ\); the validation and
calibration pools contain three rotations each. The three sets of angles are sampled from separate random streams for fitting,
checkpoint selection, and calibration. Rotation
introduces padding regions that carry no object content: those tokens are masked
out of every pool, and therefore out of the standardization statistics
\(\mu_\ell,\sigma_\ell\), the fitting batches, the validation loss, and the
calibration statistics.

\smallskip
\noindent\textbf{Standardized multi-depth tokens.}
With \(h_p^{(\ell)}(I)\in\mathbb R^d\) denoting the DINOv2 token at position \(p\) of block \(\ell\), the standardized multi-depth token \(x_p(I)\) is defined as
\begin{equation}
x_p(I)
=
\frac{1}{|\mathcal B|}
\sum_{\ell\in\mathcal B}
\frac{h_p^{(\ell)}(I)-\mu_\ell}{\sigma_\ell+\eta},
\label{eq:supp_semantic_token}
\end{equation}
where \(\mathcal B\) is the set of selected blocks, \(\mu_\ell\in\mathbb R^d\) and \(\sigma_\ell\in\mathbb R^d\) are the channel-wise mean and standard deviation of block \(\ell\) evaluated over valid tokens of the fitting views, and \(\eta>0\) ensures numerical stability. Only the fitting views define \(\mu_\ell\) and \(\sigma_\ell\), which are then applied unchanged to validation, calibration, and query tokens.

\subsection{Architecture and Optimization}
\label{supp:architecture_optimization}
The velocity network receives \(z_t\) and a shared 256-dimensional sinusoidal
embedding of \(t\). The time embedding is concatenated with \(z_t\) before
projection to a bottleneck of width \(d/2\), and also provides the scale and
shift used to modulate each of four residual blocks. After projection back to
\(d\) dimensions, a learned scalar function of \(t\) gates an input-aligned
skip from \(z_t\). The output projection and skip gate are initialized at zero.
The resulting velocity network contains 23.2\,M trainable parameters. The
network is optimized with AdamW, weight decay
\(10^{-4}\), and a cosine learning-rate schedule from \(10^{-3}\) to
\(10^{-6}\) for at most 1000 epochs. Training employs batches of 10,240 tokens,
mixed precision, and gradient-norm clipping at \(1.0\). Fixed-draw validation
is performed every three epochs; early stopping operates with a patience of ten
evaluations and requires an improvement of at least \(10^{-4}\).

\subsection{Calibration and Scoring}
\label{supp:calibration_scoring}

Operating-time calibration takes three calibration views per support image,
giving \(M=3k\) total views in the \(k\)-shot setting. These views are
independent rotations when rotation augmentation is enabled and identity views
otherwise. In the full-shot setting, \(M\) instead denotes the number of images
in the held-out calibration partition. Calibration evaluates \(J=4\)
independent Gaussian-noise repetitions and \(B=4\) random probes per token, with
the numerical-stability constant set to \(\delta=10^{-8}\). All candidate times
in \(\mathcal T=\{0.05,0.10,\ldots,0.95\}\) share the same noise draws and probe
directions, reducing sampling variation in comparisons between candidates.

\smallskip
\noindent\textbf{Definitions.}
Let \(m\in\{1,\ldots,M\}\) index calibration views, \(\mathcal P_m\) the
positions of view \(m\) whose tokens are valid, that is, outside augmentation
padding, and \(x_{m,p}^{\mathrm{cal}}\) the token at position \(p\) of view
\(m\). For each view, \(j\in\{1,\ldots,J\}\) indexes independent Gaussian draws
\(\epsilon_{m,j,p}\sim\mathcal N(0,I_d)\), one per token. With
\begin{equation}
\rho_t(x;\epsilon)
=
x-\mathcal R_{\widehat{\theta},t}\!\left(tx+(1-t)\epsilon\right)
\label{eq:supp_residual_function}
\end{equation}
denoting the residual of the main paper, where
\(\mathcal R_{\widehat\theta,t}(z)=z+(1-t)v_{\widehat\theta}(z,t)\) is the recovery
operator, and \(J_{\rho_t}(x;\epsilon)=\partial\rho_t(x;\epsilon)/\partial x\) its
Jacobian with respect to the token while holding \(\epsilon\) fixed, the two
terms of the JGN ratio are
\begin{align}
\mathcal G_{m,j}(t)
&=
\operatorname*{mean}_{p\in\mathcal P_m}
\mathbb E_u
\left\|
J_{\rho_t}\!\left(
x_{m,p}^{\mathrm{cal}};\epsilon_{m,j,p}
\right)u
\right\|_2^2 ,
\label{eq:supp_gain_term}\\
\mathcal F_{m,j}(t)
&=
\operatorname{TopMean}_{1\%}
\left\{
\left\|
\rho_t\!\left(x_{m,p}^{\mathrm{cal}};\epsilon_{m,j,p}\right)
\right\|_2^2
:p\in\mathcal P_m
\right\},
\label{eq:supp_residual_floor}
\end{align}
where the expectation in Eq.~\eqref{eq:supp_gain_term} is over random unit
directions \(u\in\mathbb R^d\). The gain term therefore averages over positions and directions, whereas the
upper-tail term aggregates only over positions to summarize the largest
residuals produced by a normal view.

\smallskip
\noindent\textbf{Why the ratio is well posed.}
Both the residual and its Jacobian contain a factor \(1-t\), so
\(\mathcal G_{m,j}(t)\) and \(\mathcal F_{m,j}(t)\) share a \((1-t)^2\) factor.
Apart from the stabilizer \(\delta\), this factor cancels in their ratio,
preventing a trivial preference induced by remaining-time scaling: a criterion
built on the gain alone would drift toward small \(t\) purely through this
scaling.

\smallskip
\noindent\textbf{Estimating the direction-averaged Jacobian gain.}
The probes \(u_{m,j,p,b}\in\mathbb R^d\), with \(b\in\{1,\ldots,B\}\), are
Rademacher vectors with entries drawn uniformly from
\(\{-1/\sqrt d,+1/\sqrt d\}\), so each probe has unit norm and
\(\mathbb E[uu^\top]=I_d/d\). For any matrix \(J\),
\[
\mathbb E_u\!\left[\lVert Ju\rVert_2^2\right]
=\mathbb E_u\!\left[\operatorname{tr}\!\left(J^\top J\,uu^\top\right)\right]
=\frac{\operatorname{tr}(J^\top J)}{d}
=\frac{\lVert J\rVert_F^2}{d},
\]
so the probe average is an unbiased estimator of the direction-averaged squared
gain, in the manner of stochastic trace estimation applied to \(J^\top J\)~\cite{hutchinson1989stochastic,avron2011randomized}. Replacing the
expectation in Eq.~\eqref{eq:supp_gain_term} by its \(B\)-probe estimate gives the
quantity actually computed,
\[
\widehat{\mathcal G}_{m,j}(t)
=
\operatorname*{mean}_{\substack{p\in\mathcal P_m\\ b\in\{1,\ldots,B\}}}
\left\lVert
J_{\rho_t}\!\left(
x_{m,p}^{\mathrm{cal}};\epsilon_{m,j,p}
\right)u_{m,j,p,b}
\right\rVert_2^2 ,
\]
where each Jacobian--vector product is obtained by one forward-mode
automatic-differentiation pass through the residual, so the full Jacobian is
never materialized. The finite-probe estimate
\(\widehat{\mathcal G}_{m,j}(t)\) is substituted for
\(\mathcal G_{m,j}(t)\) when computing JGN.
JGN is computed through the complete recovery mapping,
including the time-dependent input skip. Rademacher probes are adopted because anomaly directions are
unavailable under normal-only calibration, so isotropic unit probes evaluate
the direction-averaged response by construction.

\smallskip
\noindent\textbf{Operating-time selection.}
The final operating time \(t^\star\) selects the earliest candidate maximizing the median of the JGN ratio \(\mathcal C_{m,j}(t)=\mathcal G_{m,j}(t)/(\mathcal F_{m,j}(t)+\delta)\) across noise draws and calibration views:
\begin{equation}
t^\star
=
\min\!\left(
\operatorname*{arg\,max}_{t\in\mathcal T}
\left[
\operatorname*{median}_{m}
\operatorname*{median}_{j}
\mathcal C_{m,j}(t)
\right]
\right).
\label{eq:supp_operating_time_selection}
\end{equation}

At inference, the residual map is Gaussian-smoothed with \(\sigma=0.8\) token widths before bilinear upsampling.
Image scoring aggregates the unsmoothed upsampled map.

\subsection{Runtime, Memory, and Model Size}
\label{supp:efficiency_details}
All measurements are conducted on a single NVIDIA H100 GPU with mixed precision, reported as category-wise medians for a single seed (spread is across categories). Latency is measured from preprocessed input tensors with the first batch discarded as warm-up. In the 1-shot setting, velocity-field fitting takes a median of 24\,s per category on MVTec-AD and 28\,s on VisA (maxima 37 and 72\,s, scaling linearly with \(k\)), while JGN calibration adds 14\,s; feature extraction takes 2\,s. Peak GPU memory across all stages is 2.6\,GB. The velocity network contains 23.2\,M parameters (\(\approx\)88\,MiB), while the frozen 1.1\,B-parameter encoder is shared across categories; unlike token memory banks, the velocity network's footprint does not grow with \(k\).

\renewcommand{\topfraction}{0.95}
\renewcommand{\bottomfraction}{0.95}
\renewcommand{\textfraction}{0.05}
\renewcommand{\floatpagefraction}{0.8}
\renewcommand{\dbltopfraction}{0.95}
\renewcommand{\dblfloatpagefraction}{0.8}

\section{Additional Analyses}
\label{supp:additional_analyses}

\subsection{Recovery Contracts Toward Normal Support Features}
\label{supp:recovery_mechanism}

For the population-optimal field, single-step recovery satisfies
\[
\mathbb E[X\mid Z_t=z]=z+(1-t)v^\star(z,t).
\]
This identity motivates normality-biased recovery, but does not establish how a
field learned from finite normal supports behaves on anomalous queries. This
analysis examines the implied mechanism directly: tokens farther from the empirical
normal set should undergo larger recovery changes, and those changes should
move them toward that set.

Let \(\mathcal X_{\mathrm{sup}}\) denote the token bank extracted from the
unrotated normal support images assigned to fitting. This finite bank serves
only as an empirical reference for the following analysis; it excludes the
rotation-augmented fitting views and should not be interpreted as the complete
training distribution or an underlying normal manifold. Throughout this
subsection, \(q\) abbreviates a query token \(x_p^q\) of the main paper and
\(\widehat{x}(q)\) its recovered counterpart. For each query token, its
nearest-support distance is evaluated before and after recovery:
\begin{equation}
\begin{aligned}
d_{\mathrm{before}}(q)
&=
\min_{x\in\mathcal X_{\mathrm{sup}}}\|q-x\|_2,\\
d_{\mathrm{after}}(q)
&=
\min_{x\in\mathcal X_{\mathrm{sup}}}
\|\widehat{x}(q)-x\|_2.
\end{aligned}
\label{eq:support_distances}
\end{equation}
These are distances to the finite support set, not to an assumed underlying
normal manifold. Let
\[
x_{\mathrm{NN}}(q)
=
\operatorname*{arg\,min}_{x\in\mathcal X_{\mathrm{sup}}}\|q-x\|_2.
\]
Recovery dynamics are characterized via
\begin{equation}
\begin{aligned}
r(q)
&=
\|q-\widehat{x}(q)\|_2^2,\\
\gamma(q)
&=
\frac{d_{\mathrm{after}}(q)}
     {d_{\mathrm{before}}(q)},\\
a(q)
&=
\frac{
\bigl(\widehat{x}(q)-q\bigr)^\top
\bigl(x_{\mathrm{NN}}(q)-q\bigr)
}{
\|\widehat{x}(q)-q\|_2
\|x_{\mathrm{NN}}(q)-q\|_2
}.
\end{aligned}
\label{eq:support_diagnostics}
\end{equation}
The correlation between \(r(q)\) and \(d_{\mathrm{before}}(q)\) measures
whether distance from normal support is reflected in the anomaly score.
A ratio \(\gamma(q)<1\) indicates contraction toward the support set, while
\(a(q)>0\) indicates that the recovery update has a component pointing toward
the nearest support token.

Statistics are computed separately within each category and then summarized
across categories. This prevents categories with more test tokens from
dominating the analysis. For the binned curves, distances and residuals are
normalized by their category-wise normal-token medians. Ground-truth masks
serve only for this post-hoc stratification of normal and anomalous tokens.
\begin{figure*}[t]
\centering
\includegraphics[width=\textwidth]
{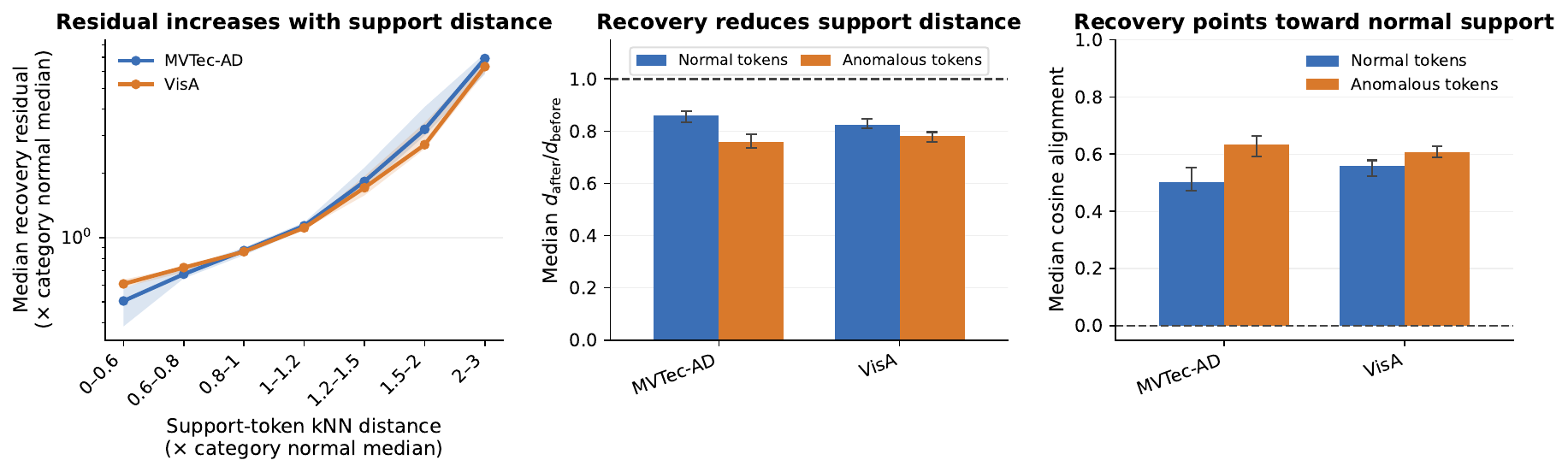}
\caption{
Recovery relative to the support-derived normal token set. Left: recovery
residual versus pre-recovery support distance, with both quantities normalized
by their category-wise normal-token medians and aggregated equally across
categories. Middle: \(d_{\mathrm{after}}/d_{\mathrm{before}}\), where values
below one indicate contraction toward the normal support features. Right:
cosine alignment between the recovery update and the direction toward the
nearest support token, where positive values indicate movement toward the
normal support features. Bars and error bars report
the median and interquartile range across categories. Ground-truth masks
serve only for post-hoc stratification.
}
\label{fig:supp_recovery_mechanism}
\end{figure*}

Figure~\ref{fig:supp_recovery_mechanism} reveals the same pattern on both
datasets. The recovery residual increases monotonically across
nearest-support-distance bins, with median category-wise Spearman correlations
of \(0.825\) on MVTec-AD and \(0.817\) on VisA. The association remains strong
within normal tokens (\(0.803/0.809\)) and anomalous tokens
(\(0.921/0.918\)), rather than arising only from differences between the two
groups.

Recovered counterparts generally contract toward the normal support features.
The median
\(d_{\mathrm{after}}/d_{\mathrm{before}}\) is \(0.861/0.826\) for normal
tokens and \(0.760/0.782\) for anomalous tokens on MVTec-AD/VisA. Relative to
each category's normal-token distance scale, the absolute contraction is
\(2.8\times/2.2\times\) larger for anomalous tokens. Recovery is directionally
consistent with this contraction: median cosine alignment with the
nearest-support direction is \(0.503/0.558\) for normal tokens and
\(0.635/0.605\) for anomalous tokens.

Therefore, larger residuals are associated with greater support distance, while
recovery updates are generally directed toward empirical normal support features.
Although this does not constitute an anomaly-separation guarantee, the
observed behavior is consistent with the normality-biased recovery mechanism
underlying Flownomaly.

\subsection{Operating-Time Selection and Robustness}
\label{supp:time_calibration}

JGN calibrates a category-specific operating time \(t^\star\) from support-derived normal calibration views alone. To assess selection quality, retrospective regret is evaluated relative to an optimistic test-label oracle (the candidate time maximizing test image AUROC, evaluated strictly post hoc).

As reported in Table~\ref{tab:supp_time_regret}, median regret is only \(0.06\)
percentage points (pp) on MVTec-AD and \(0.14\) on VisA. VisA's larger mean and
maximum (4.82\,pp) are driven by \textit{macaroni2}, whose empirical test curve
favors an earlier operating time. These regret statistics, together with the
category-wise diagnostics in Table~\ref{tab:supp_time_categories}, are computed
for seed 42; the per-category median selection for this sampling is \(0.65\) on
both datasets.

\begin{figure}[!ht]
\centering
\includegraphics[width=0.95\linewidth]{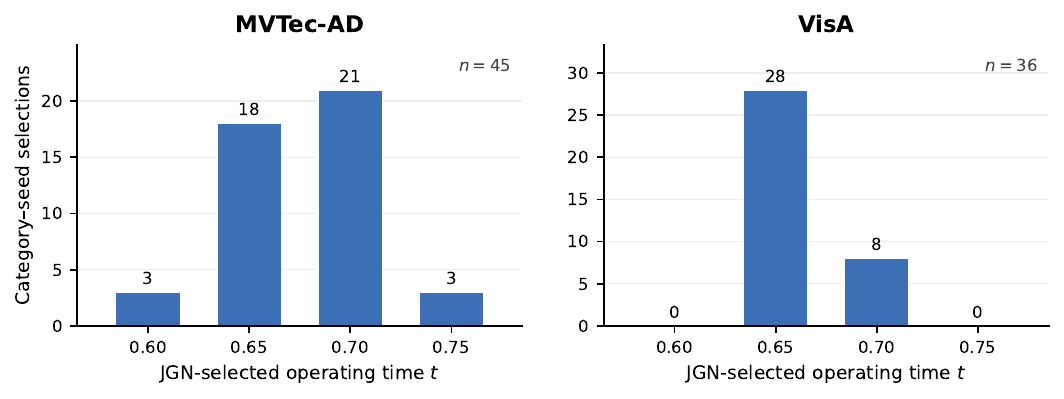}
\vspace{-2pt}
\caption{Distribution of category-specific operating times selected by JGN,
pooled across three seeds. Each observation represents one category--seed pair.}
\label{fig:supp_selection_frequencies}
\end{figure}

Figure~\ref{fig:supp_selection_frequencies} pools all category--seed pairs from
seeds 42--44. Selections remain confined to \(t\in[0.60,0.75]\) on MVTec-AD and
\([0.65,0.70]\) on VisA. Pooled
over all 45 MVTec-AD pairs, the \(0.65\) and \(0.70\) bins are near-evenly
populated (18 and 21 selections); on VisA, \(0.65\) accounts for 28 of 36 pairs.

\begin{table}[!ht]
\centering
\caption{
Retrospective image-AUROC regret of the normal-only JGN selection, measured in percentage points relative to the category-wise test-label oracle. Regret
statistics are computed over categories for seed 42; the \(t^\star\) column reports the range of per-category selections
across all three seeds.
}
\label{tab:supp_time_regret}
\vspace{2pt}
\setlength{\tabcolsep}{4.8pt}%
\begin{tabular}{lc|ccc}
\toprule
& & \multicolumn{3}{c}{Regret (pp)} \\
\cmidrule(lr){3-5}
Dataset & \(t^\star\) range & Mean & Median & Maximum \\
\midrule
MVTec-AD & \(0.60\)--\(0.75\) & 0.15 & 0.06 & 0.82 \\
VisA & \(0.65\)--\(0.70\) & 0.54 & 0.14 & 4.82 \\
\bottomrule
\end{tabular}
\end{table}

\begin{table}[!ht]
\centering
\caption{
Category-level operating-time diagnostics for seed 42. Regret is the
image-AUROC difference, in percentage points, from the retrospective
test-label oracle.
}
\label{tab:supp_time_categories}
\vspace{2pt}
\resizebox{\columnwidth}{!}{%
\renewcommand{\arraystretch}{0.92}%
\setlength{\tabcolsep}{3pt}%
\begin{tabular}{lrr@{\hspace{1.2em}}lrr}
\toprule
\multicolumn{3}{c}{MVTec-AD} & \multicolumn{3}{c}{VisA} \\
\cmidrule(lr){1-3}\cmidrule(lr){4-6}
Category & \(t^\star\) & Regret (pp) & Category & \(t^\star\) & Regret (pp) \\
\midrule
bottle & 0.75 & 0.08 & candle & 0.70 & 0.09 \\
cable & 0.65 & 0.82 & capsules & 0.65 & 0.36 \\
capsule & 0.65 & 0.08 & cashew & 0.65 & 0.02 \\
carpet & 0.70 & 0.00 & chewinggum & 0.65 & 0.04 \\
grid & 0.75 & 0.00 & fryum & 0.65 & 0.23 \\
hazelnut & 0.70 & 0.00 & macaroni1 & 0.65 & 0.26 \\
leather & 0.70 & 0.00 & macaroni2 & 0.70 & 4.82 \\
metal\_nut & 0.65 & 0.00 & pcb1 & 0.65 & 0.18 \\
pill & 0.70 & 0.61 & pcb2 & 0.65 & 0.03 \\
screw & 0.65 & 0.23 & pcb3 & 0.65 & 0.00 \\
tile & 0.70 & 0.00 & pcb4 & 0.65 & 0.02 \\
toothbrush & 0.65 & 0.14 & pipe\_fryum & 0.65 & 0.45 \\
transistor & 0.60 & 0.06 & \textemdash{} & \textemdash{} & \textemdash{} \\
wood & 0.65 & 0.22 & \textemdash{} & \textemdash{} & \textemdash{} \\
zipper & 0.65 & 0.00 & \textemdash{} & \textemdash{} & \textemdash{} \\
\bottomrule
\end{tabular}%
}
\end{table}

\subsection{Feature and Support-View Ablations}
\label{supp:feature_ablations}

As shown in Table~\ref{tab:design_ablation}, feature depth has the clearest and most consistent effect. Replacing the default 40--80\% depth window with shallow 20--40\% blocks reduces image AUROC by 12.1~pp on MVTec-AD and 9.3~pp on VisA, while pixel AUROC decreases by 2.1 and 5.2~pp. Retaining only the block at 80\% depth preserves MVTec-AD performance but lowers VisA image AUROC by 3.3~pp.

\begin{table*}[t]
\centering
\caption{Feature and support-view ablations in the 1-shot setting. AUROC
results are mean \(\pm\) standard deviation over three seeds. The final column reports the median selected \(t^\star\) over categories for MVTec-AD/VisA under seed 42. \(^\dagger\)Disabling rotations also removes diversity from the support-derived validation and calibration pools.}
\label{tab:design_ablation}
\begin{tabular}{l|cc|cc|c}
\toprule
& \multicolumn{2}{c|}{MVTec-AD} & \multicolumn{2}{c}{VisA} & \\
Configuration & I-AUROC & P-AUROC & I-AUROC & P-AUROC & Median \(t^*\) (MVTec-AD/VisA) \\
\midrule
Default: 40--80\%, standardized
& \resultpm{97.8}{0.9} & \resultpm{98.0}{0.1}
& \resultpm{95.3}{0.7} & \resultpm{98.7}{0.0} & 0.65/0.65 \\
Single block at 80\% depth
& \resultpm{97.6}{1.2} & \resultpm{98.0}{0.0}
& \resultpm{92.0}{0.7} & \resultpm{98.9}{0.0} & 0.70/0.70 \\
Shallow blocks at 20--40\% depth
& \resultpm{85.7}{0.9} & \resultpm{95.9}{0.1}
& \resultpm{86.0}{1.7} & \resultpm{93.5}{0.3} & 0.60/0.60 \\
Without layer standardization
& \resultpm{97.7}{1.3} & \resultpm{98.2}{0.1}
& \resultpm{94.0}{0.9} & \resultpm{98.8}{0.0} & 0.85/0.85 \\
Without rotation-derived views\(^\dagger\)
& \resultpm{93.8}{1.4} & \resultpm{97.4}{0.1}
& \resultpm{93.3}{0.5} & \resultpm{98.5}{0.0} & 0.60/0.60 \\
\bottomrule
\end{tabular}
\end{table*}

Removing layer-wise standardization has little effect on MVTec-AD but reduces
VisA image AUROC by 1.3~pp. Removing rotation-derived views reduces image
AUROC by 4.0~pp on MVTec-AD and 2.0~pp on VisA. This ablation
also removes view diversity from the validation and calibration pools, therefore measuring the complete support-view construction rather than only fitting.

The JGN-selected operating time also changes with the feature representation:
the median \(t^\star\) shifts from \(0.60\) for shallow features to \(0.85\)
without layer standardization, with the same ordering on both datasets,
showing that the selected operating time responds to changes in the feature
representation rather than remaining fixed.

\subsection{Qualitative Failure Modes}
\label{supp:failure_modes}

Flownomaly scores token-wise feature discrepancies against the support-derived normal distribution. While effective for localized defects, it does not explicitly model object-level relations, part counts, or global configurations.

\begin{figure}[!ht]
\centering
\includegraphics[width=\linewidth]
{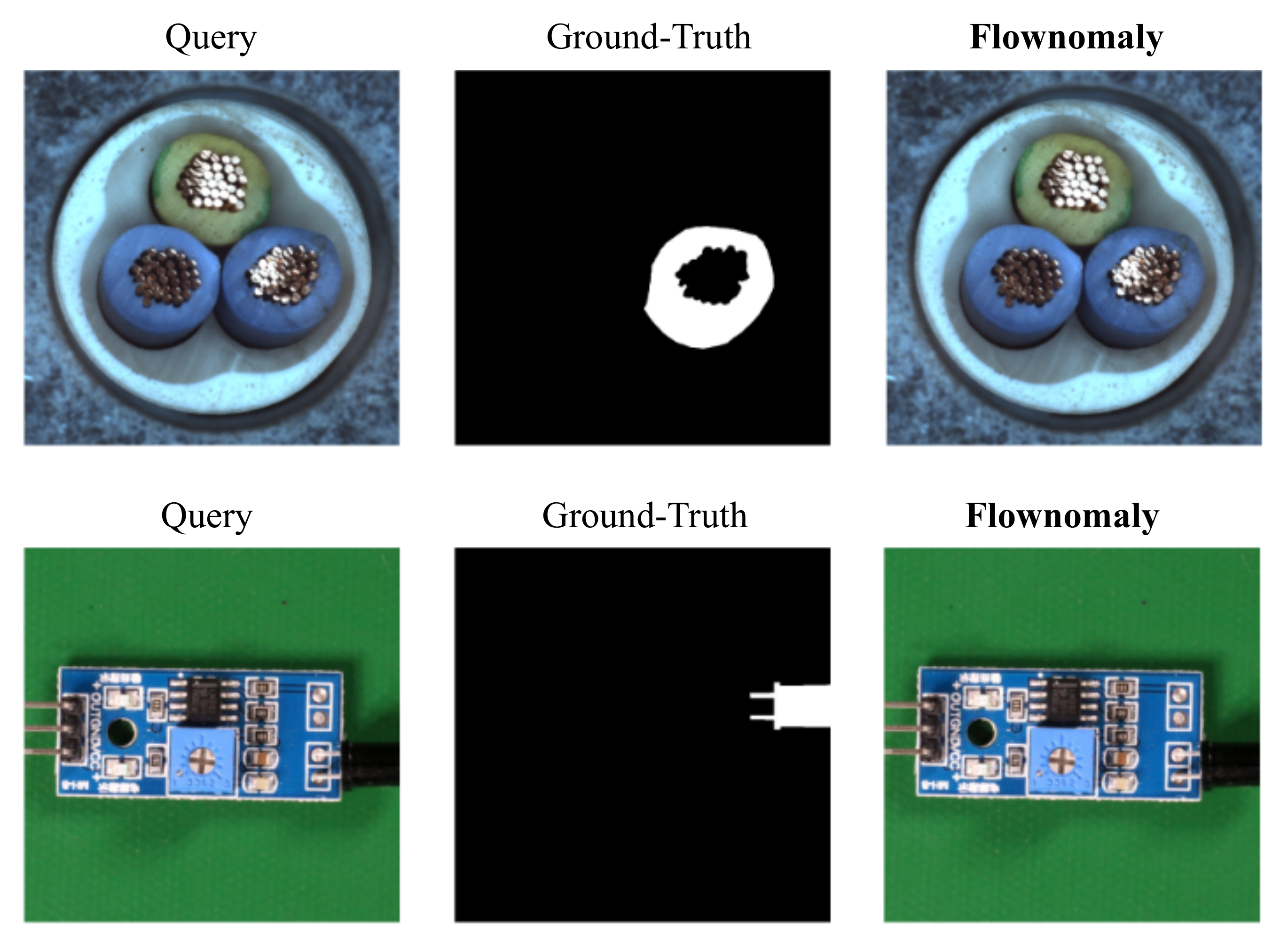}
\caption{
Failure modes where the predicted anomaly map remains essentially empty.
Columns show the query image, ground-truth mask, and Flownomaly anomaly map.
Top: Logical anomaly with swapped wires (MVTec-AD \textit{cable}).
Bottom: Missing component revealing previously seen background (VisA \textit{pcb3}).
}
\label{fig:supp_failure_modes}
\end{figure}

Figure~\ref{fig:supp_failure_modes} illustrates two representative failure cases where the predicted anomaly map remains empty:
\begin{itemize}[leftmargin=*,noitemsep,topsep=2pt]
\item \textbf{Logical anomalies:} In MVTec-AD \textit{cable} with swapped wires (top), individual wire segments appear locally normal, producing near-zero residuals and an empty anomaly map despite the invalid configuration.
\item \textbf{Missing components:} In VisA \textit{pcb3} (bottom), the missing component reveals the same background as previously seen in normal support views; because the observed tokens remain compatible with the normal distribution, the predicted anomaly map is blank.
\end{itemize}

\section{Additional Evaluations}
\label{supp:additional_evaluations}

\subsection{Backbone Comparison}
\label{supp:additional_backbones}

Table~\ref{tab:dinov3_vs_dinov2_ablation} complements the Table 5 of the main paper by comparing against DINOv3-7B~\cite{simeoni2025dinov3}. DINOv3 employs a \(16\times16\) patch size (versus DINOv2's \(14\times14\)), so the default \(672\times672\) input resolution produces a \(42\times42\) patch token grid. To ensure a direct comparison on an identical feature map resolution, DINOv3 is additionally evaluated at a higher resolution of \(768\times768\), matching the default \(48\times48\) grid (\(768/16=48\)). As shown in Table~\ref{tab:dinov3_vs_dinov2_ablation}, DINOv3-7B does not improve over DINOv2-g at either resolution, consistent with findings reported by SubspaceAD~\cite{subspacead}. As in the DINOv2 ablation, the velocity-network bottleneck width is half the encoder's token embedding dimension (\(d/2\)).

\begin{table}[!ht]
\centering
\caption{DINOv3 vs. DINOv2 backbone ablation in the 1-shot setting across MVTec-AD and VisA. DINOv3 (\(16\times16\) patch size) is evaluated at \(672\times672\) (\(42\times42\) token grid) and at \(768\times768\) to match DINOv2's \(48\times48\) token grid.}
\label{tab:dinov3_vs_dinov2_ablation}
\resizebox{\columnwidth}{!}{%
\setlength{\tabcolsep}{5.0pt}%
\begin{tabular}{lc|cc|cc}
\toprule
& & \multicolumn{2}{c|}{MVTec-AD} & \multicolumn{2}{c}{VisA} \\
Backbone & Res. & I-AUROC & P-AUROC & I-AUROC & P-AUROC \\
\midrule
DINOv2-g & \(672^2\) & \resultpm{97.8}{0.9} & \resultpm{98.0}{0.1} & \resultpm{95.3}{0.7} & \resultpm{98.7}{0.0} \\
DINOv3-7B & \(672^2\) & \resultpm{96.1}{1.0} & \resultpm{98.0}{0.1} & \resultpm{94.8}{0.1} & \resultpm{98.5}{0.0} \\
DINOv3-7B & \(768^2\) & \resultpm{96.0}{1.6} & \resultpm{97.9}{0.1} & \resultpm{94.9}{0.2} & \resultpm{98.5}{0.0} \\
\bottomrule
\end{tabular}%
}
\end{table}

\subsection{Full-Shot Evaluation}
\label{supp:full_shot}

Although Flownomaly is designed for FSAD, performance is also evaluated across all
available normal training images. Rotations are disabled, and the images are
partitioned into fitting, validation, and calibration pools.
For \(n\) normal training images, validation and calibration each receive
\(\max(4,\lceil0.05n\rceil)\) images.
Tables~\ref{tab:full_shot_mvtec} and~\ref{tab:full_shot_visa} report the aggregate results on MVTec-AD and VisA, respectively.

\begin{table}[!ht]
\centering
\caption{
    Full-shot anomaly detection and localization on MVTec-AD. Rotation augmentations are disabled in the full-shot setting.
    All values are percentages.
    \textbf{Bold} denotes the best result, and
    \underline{underlined} denotes the second-best result.
}
\label{tab:full_shot_mvtec}
\resizebox{\columnwidth}{!}{%
\renewcommand{\arraystretch}{1.08}%
\setlength{\tabcolsep}{5pt}%
\begin{tabular}{l|ccc|ccc}
\toprule
& \multicolumn{3}{c|}{Image-level} & \multicolumn{3}{c}{Pixel-level} \\
Method & AUROC & AUPR & \(F_{1,\max}\) & AUROC & AUPRO & \(F_{1,\max}\) \\
\midrule
PatchCore~\cite{patchcore}     & 99.2 & \underline{99.8} & \underline{98.8} & 98.1 & 92.6 & 59.3 \\
AnomalyDINO~\cite{anomalydino} & \underline{99.5} & \underline{99.8} & \textbf{99.0} & \underline{98.2} & 95.0 & 64.3 \\
SubspaceAD~\cite{subspacead}   & 99.2 & 99.6 & 98.3 & \underline{98.2} & \underline{95.6} & 64.7 \\
ANoCo~\cite{anoco}             & 99.4 & 99.7 & 98.5 & \underline{98.2} & \textbf{96.4} & \textbf{67.8} \\
\midrule
\textbf{Flownomaly}            & \textbf{99.7} & \textbf{99.9} & \textbf{99.0} & \textbf{98.7} & \textbf{96.4} & \underline{67.2} \\
\bottomrule
\end{tabular}%
}
\end{table}

\begin{table}[!ht]
\centering
\caption{
    Full-shot anomaly detection and localization on VisA. Rotation augmentations are disabled in the full-shot setting.
    All values are percentages.
    \textbf{Bold} denotes the best result, and
    \underline{underlined} denotes the second-best result.
}
\label{tab:full_shot_visa}
\resizebox{\columnwidth}{!}{%
\renewcommand{\arraystretch}{1.08}%
\setlength{\tabcolsep}{5pt}%
\begin{tabular}{l|ccc|ccc}
\toprule
& \multicolumn{3}{c|}{Image-level} & \multicolumn{3}{c}{Pixel-level} \\
Method & AUROC & AUPR & \(F_{1,\max}\) & AUROC & AUPRO & \(F_{1,\max}\) \\
\midrule
PatchCore~\cite{patchcore}     & 92.8 & 91.4 & 88.0 & 98.5 & 91.1 & 43.7 \\
AnomalyDINO~\cite{anomalydino} & 97.6 & 98.0 & 94.5 & 98.8 & 96.1 & 53.8 \\
SubspaceAD~\cite{subspacead}   & 98.2 & \underline{98.3} & 95.0 & 99.1 & \underline{96.9} & 53.0 \\
ANoCo~\cite{anoco}             & \underline{98.7} & \textbf{98.9} & \underline{96.1} & \underline{99.2} & 95.3 & \underline{54.7} \\
\midrule
\textbf{Flownomaly}            & \textbf{98.9} & \textbf{98.9} & \textbf{96.5} & \textbf{99.3} & \textbf{97.0} & \textbf{54.8} \\
\bottomrule
\end{tabular}%
}
\end{table}

\section{Additional Qualitative Results}
\label{supp:qualitative_results}

\begin{figure*}[!t]
\centering
\includegraphics[width=\linewidth,height=0.88\textheight,keepaspectratio]{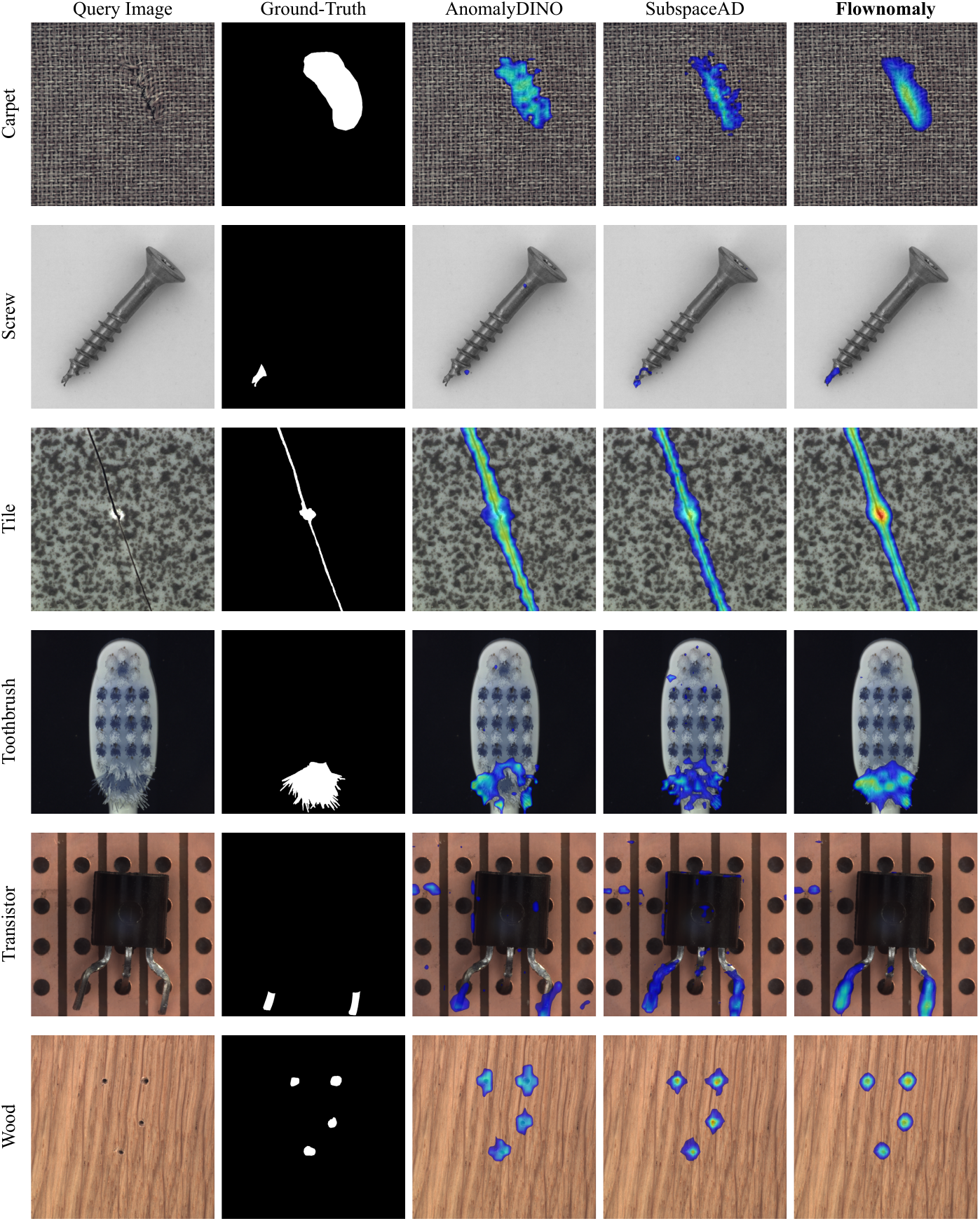}
\caption{
Additional qualitative localization results on representative MVTec-AD samples under the 1-shot setting.
Columns show the input image, ground-truth mask, baseline anomaly maps (AnomalyDINO~\cite{anomalydino}, SubspaceAD~\cite{subspacead}), and Flownomaly's predicted anomaly map.
}
\label{fig:supp_qualitative_mvtec}
\end{figure*}

\begin{figure*}[!t]
\centering
\includegraphics[width=\linewidth,height=0.88\textheight,keepaspectratio]{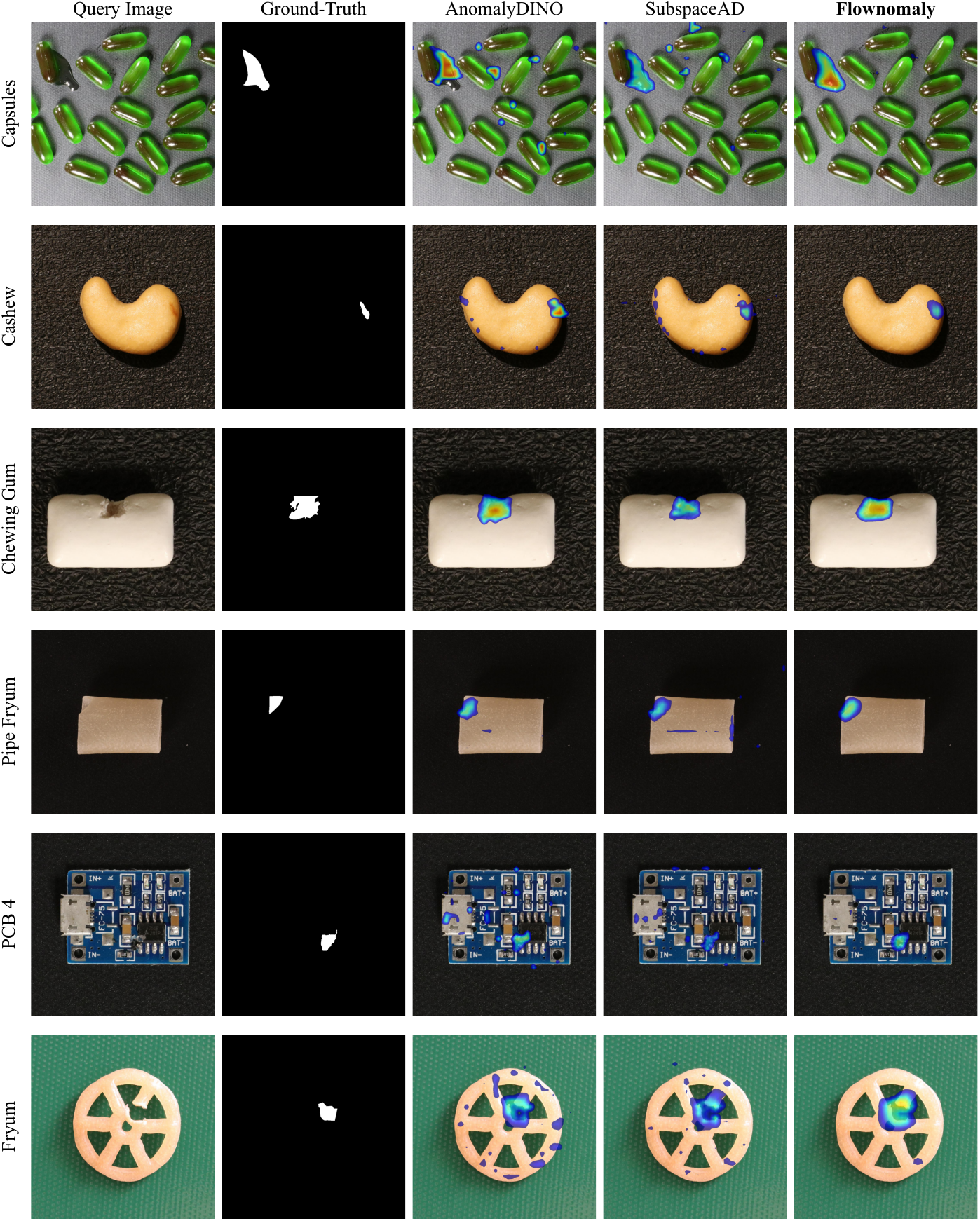}
\caption{
Additional qualitative localization results on representative VisA samples under the 1-shot setting.
Columns show the input image, ground-truth mask, baseline anomaly predictions (AnomalyDINO~\cite{anomalydino}, SubspaceAD~\cite{subspacead}), and Flownomaly.
}
\label{fig:supp_qualitative_visa}
\end{figure*}

\begin{figure*}[!t]
\centering
\includegraphics[width=\linewidth,height=0.88\textheight,keepaspectratio]{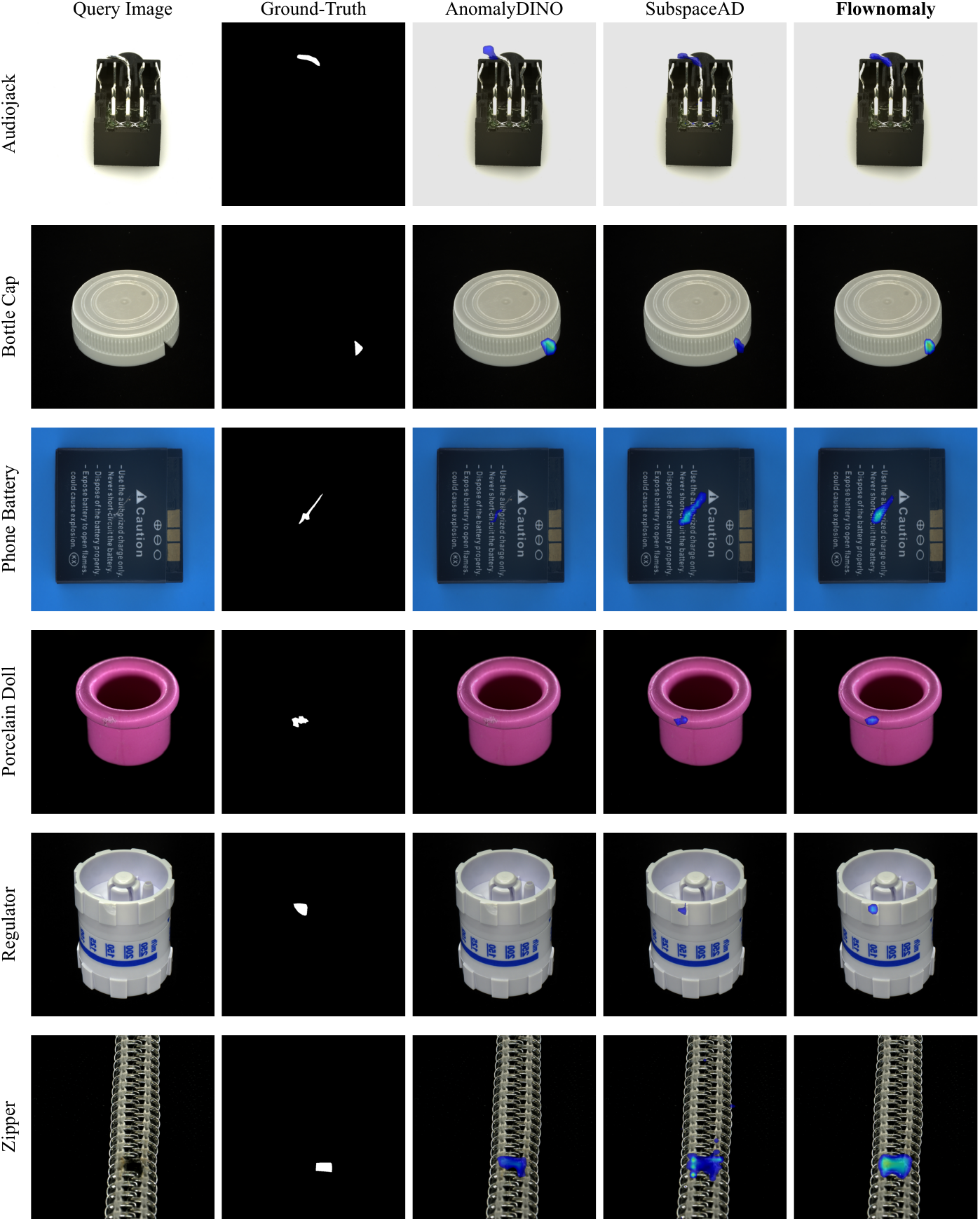}
\caption{
Qualitative localization results on representative Real-IAD samples under the 1-sample setting (spanning five multi-view captures of a single normal physical sample).
Columns show the input query image, ground-truth mask, baseline anomaly maps (AnomalyDINO~\cite{anomalydino}, SubspaceAD~\cite{subspacead}), and Flownomaly.
}
\label{fig:supp_qualitative_realiad}
\end{figure*}

To complement the qualitative results in the main paper, this section provides an extended set of qualitative comparisons, presenting detailed anomaly maps for representative samples from the MVTec-AD dataset in Fig.~\ref{fig:supp_qualitative_mvtec}, the VisA dataset in 
Fig.~\ref{fig:supp_qualitative_visa}, and the Real-IAD dataset in Fig.~\ref{fig:supp_qualitative_realiad}. These examples further 
illustrate the model's localization performance across diverse object categories and anomaly types under the 1-shot (and 1-sample) 
settings, confirming that Flownomaly consistently achieves sharp defect delineation while suppressing background responses.

\clearpage
\onecolumn

\section{Per-Category Few-Shot and Sample-Shot Results}
\label{supp:per_category_results}

Tables~\ref{tab:supp_per_object_mvtec} and~\ref{tab:supp_per_object_visa} report
per-category results on MVTec-AD and VisA under the 1-, 2-, and 4-shot settings. For Real-IAD, due to the 30-category scale, per-category results are partitioned across two separate tables: Table~\ref{tab:supp_per_object_realiad_wide_1_2} reports the 1- and 2-sample settings, and Table~\ref{tab:supp_per_object_realiad_wide_4} reports the 4-sample setting.

\begin{center}
\captionof{table}{
    Per-category anomaly detection and localization results on MVTec-AD under the
    1-, 2-, and 4-shot settings.
    Results are mean and sample standard deviation over three seeds.
    All values are percentages.
}
\label{tab:supp_per_object_mvtec}

\vspace{2pt}

\small
\renewcommand{\arraystretch}{0.96}%
\setlength{\tabcolsep}{5pt}%
\begin{tabular}{c l|ccc|ccc}
\toprule

\multirow{3}{*}{\textbf{Setup}}
& \multirow{3}{*}{\textbf{Category}}
& \multicolumn{6}{c}{\textbf{MVTec-AD}}
\\

&
& \multicolumn{3}{c|}{\textbf{Image-level}}
& \multicolumn{3}{c}{\textbf{Pixel-level}}
\\

&
& AUROC & AUPR & \(F_{1,\max}\)
& AUROC & AUPRO & \(F_{1,\max}\)
\\

\midrule

% =========================================================
% 1-shot
% =========================================================

\multirow{15}{*}{\textbf{1-shot}}
& bottle & \resultpm{99.9}{0.0} & \resultpm{100.0}{0.0} & \resultpm{99.2}{0.0} & \resultpm{98.9}{0.1} & \resultpm{97.5}{0.1} & \resultpm{78.2}{0.7} \\

& cable & \resultpm{94.3}{0.6} & \resultpm{97.2}{0.3} & \resultpm{92.2}{0.6} & \resultpm{96.2}{0.2} & \resultpm{91.6}{0.5} & \resultpm{66.8}{2.3} \\

& capsule & \resultpm{88.1}{13.7} & \resultpm{96.7}{4.2} & \resultpm{95.2}{2.3} & \resultpm{98.4}{0.2} & \resultpm{97.6}{0.8} & \resultpm{50.7}{8.7} \\

& carpet & \resultpm{100.0}{0.0} & \resultpm{100.0}{0.0} & \resultpm{100.0}{0.0} & \resultpm{99.5}{0.0} & \resultpm{98.6}{0.0} & \resultpm{70.4}{0.4} \\

& grid & \resultpm{100.0}{0.0} & \resultpm{100.0}{0.0} & \resultpm{100.0}{0.0} & \resultpm{99.6}{0.0} & \resultpm{98.2}{0.1} & \resultpm{59.0}{1.1} \\

& hazelnut & \resultpm{99.5}{0.8} & \resultpm{99.7}{0.4} & \resultpm{98.2}{2.6} & \resultpm{99.6}{0.1} & \resultpm{96.7}{1.3} & \resultpm{78.2}{1.5} \\

& leather & \resultpm{100.0}{0.0} & \resultpm{100.0}{0.0} & \resultpm{100.0}{0.0} & \resultpm{99.3}{0.0} & \resultpm{98.8}{0.0} & \resultpm{43.2}{0.9} \\

& metal\_nut & \resultpm{100.0}{0.0} & \resultpm{100.0}{0.0} & \resultpm{100.0}{0.0} & \resultpm{97.9}{0.2} & \resultpm{96.7}{0.2} & \resultpm{82.8}{1.8} \\

& pill & \resultpm{97.6}{0.2} & \resultpm{99.6}{0.0} & \resultpm{97.3}{0.2} & \resultpm{96.1}{0.6} & \resultpm{98.1}{0.2} & \resultpm{57.5}{1.6} \\

& screw & \resultpm{91.4}{0.3} & \resultpm{97.0}{0.1} & \resultpm{91.3}{0.3} & \resultpm{99.4}{0.0} & \resultpm{97.2}{0.1} & \resultpm{60.5}{1.3} \\

& tile & \resultpm{100.0}{0.0} & \resultpm{100.0}{0.0} & \resultpm{100.0}{0.0} & \resultpm{98.2}{0.1} & \resultpm{95.6}{0.1} & \resultpm{76.2}{0.5} \\

& toothbrush & \resultpm{98.6}{2.4} & \resultpm{99.4}{0.9} & \resultpm{97.4}{3.2} & \resultpm{99.5}{0.2} & \resultpm{97.4}{0.4} & \resultpm{70.0}{6.7} \\

& transistor & \resultpm{97.9}{0.2} & \resultpm{97.1}{0.5} & \resultpm{91.7}{1.6} & \resultpm{91.3}{1.8} & \resultpm{73.9}{2.9} & \resultpm{46.5}{3.4} \\

& wood & \resultpm{99.9}{0.2} & \resultpm{100.0}{0.1} & \resultpm{99.5}{0.9} & \resultpm{97.5}{0.3} & \resultpm{96.8}{0.1} & \resultpm{68.7}{0.1} \\

& zipper & \resultpm{99.8}{0.0} & \resultpm{99.9}{0.0} & \resultpm{98.6}{0.2} & \resultpm{99.3}{0.1} & \resultpm{98.0}{0.2} & \resultpm{70.4}{1.0} \\

\midrule

% =========================================================
% 2-shot
% =========================================================

\multirow{15}{*}{\textbf{2-shot}}
& bottle & \resultpm{100.0}{0.0} & \resultpm{100.0}{0.0} & \resultpm{99.7}{0.5} & \resultpm{99.0}{0.0} & \resultpm{97.5}{0.1} & \resultpm{78.9}{0.2} \\

& cable & \resultpm{94.6}{0.9} & \resultpm{97.4}{0.3} & \resultpm{92.8}{0.2} & \resultpm{96.5}{0.3} & \resultpm{92.4}{0.4} & \resultpm{70.3}{0.1} \\

& capsule & \resultpm{92.9}{5.5} & \resultpm{98.4}{1.4} & \resultpm{95.0}{2.3} & \resultpm{98.6}{0.0} & \resultpm{98.0}{0.1} & \resultpm{53.7}{3.5} \\

& carpet & \resultpm{100.0}{0.0} & \resultpm{100.0}{0.0} & \resultpm{100.0}{0.0} & \resultpm{99.5}{0.0} & \resultpm{98.6}{0.0} & \resultpm{70.2}{0.4} \\

& grid & \resultpm{100.0}{0.0} & \resultpm{100.0}{0.0} & \resultpm{100.0}{0.0} & \resultpm{99.6}{0.0} & \resultpm{98.3}{0.1} & \resultpm{59.4}{1.1} \\

& hazelnut & \resultpm{100.0}{0.0} & \resultpm{100.0}{0.0} & \resultpm{99.5}{0.4} & \resultpm{99.6}{0.0} & \resultpm{97.2}{0.6} & \resultpm{78.7}{0.8} \\

& leather & \resultpm{100.0}{0.0} & \resultpm{100.0}{0.0} & \resultpm{100.0}{0.0} & \resultpm{99.3}{0.0} & \resultpm{98.8}{0.1} & \resultpm{43.5}{1.0} \\

& metal\_nut & \resultpm{100.0}{0.0} & \resultpm{100.0}{0.0} & \resultpm{100.0}{0.0} & \resultpm{98.1}{0.2} & \resultpm{97.0}{0.2} & \resultpm{84.2}{1.5} \\

& pill & \resultpm{97.9}{0.4} & \resultpm{99.6}{0.1} & \resultpm{97.5}{0.0} & \resultpm{96.3}{0.3} & \resultpm{98.1}{0.1} & \resultpm{58.0}{0.8} \\

& screw & \resultpm{93.7}{1.6} & \resultpm{97.8}{0.5} & \resultpm{92.4}{1.3} & \resultpm{99.5}{0.0} & \resultpm{97.7}{0.1} & \resultpm{60.8}{1.4} \\

& tile & \resultpm{100.0}{0.0} & \resultpm{100.0}{0.0} & \resultpm{100.0}{0.0} & \resultpm{98.2}{0.1} & \resultpm{95.6}{0.2} & \resultpm{76.2}{0.6} \\

& toothbrush & \resultpm{99.4}{0.7} & \resultpm{99.8}{0.3} & \resultpm{98.3}{1.7} & \resultpm{99.4}{0.2} & \resultpm{97.4}{0.3} & \resultpm{67.9}{5.3} \\

& transistor & \resultpm{98.3}{1.5} & \resultpm{97.6}{2.2} & \resultpm{92.3}{5.3} & \resultpm{93.3}{0.8} & \resultpm{78.4}{2.5} & \resultpm{52.0}{1.1} \\

& wood & \resultpm{99.7}{0.1} & \resultpm{99.9}{0.0} & \resultpm{98.6}{0.5} & \resultpm{97.5}{0.3} & \resultpm{96.8}{0.1} & \resultpm{68.6}{0.5} \\

& zipper & \resultpm{99.9}{0.1} & \resultpm{100.0}{0.0} & \resultpm{99.0}{0.2} & \resultpm{99.3}{0.1} & \resultpm{98.0}{0.2} & \resultpm{70.3}{0.4} \\

\midrule

% =========================================================
% 4-shot
% =========================================================

\multirow{15}{*}{\textbf{4-shot}}
& bottle & \resultpm{100.0}{0.0} & \resultpm{100.0}{0.0} & \resultpm{99.7}{0.5} & \resultpm{99.1}{0.0} & \resultpm{97.5}{0.0} & \resultpm{79.6}{0.4} \\

& cable & \resultpm{94.4}{1.1} & \resultpm{97.4}{0.4} & \resultpm{92.7}{0.7} & \resultpm{96.8}{0.4} & \resultpm{93.0}{0.5} & \resultpm{71.6}{0.3} \\

& capsule & \resultpm{91.9}{6.3} & \resultpm{98.1}{1.6} & \resultpm{95.4}{2.4} & \resultpm{98.7}{0.1} & \resultpm{98.1}{0.2} & \resultpm{53.3}{3.7} \\

& carpet & \resultpm{100.0}{0.0} & \resultpm{100.0}{0.0} & \resultpm{100.0}{0.0} & \resultpm{99.4}{0.0} & \resultpm{98.5}{0.0} & \resultpm{69.6}{0.5} \\

& grid & \resultpm{100.0}{0.0} & \resultpm{100.0}{0.0} & \resultpm{100.0}{0.0} & \resultpm{99.6}{0.0} & \resultpm{98.3}{0.0} & \resultpm{59.6}{0.3} \\

& hazelnut & \resultpm{100.0}{0.0} & \resultpm{100.0}{0.0} & \resultpm{100.0}{0.0} & \resultpm{99.6}{0.0} & \resultpm{97.6}{0.3} & \resultpm{78.5}{0.5} \\

& leather & \resultpm{100.0}{0.0} & \resultpm{100.0}{0.0} & \resultpm{100.0}{0.0} & \resultpm{99.3}{0.0} & \resultpm{98.8}{0.1} & \resultpm{42.9}{1.2} \\

& metal\_nut & \resultpm{100.0}{0.0} & \resultpm{100.0}{0.0} & \resultpm{100.0}{0.0} & \resultpm{98.1}{0.1} & \resultpm{97.0}{0.2} & \resultpm{85.3}{0.6} \\

& pill & \resultpm{98.6}{0.5} & \resultpm{99.7}{0.1} & \resultpm{97.9}{0.4} & \resultpm{97.1}{0.4} & \resultpm{98.3}{0.1} & \resultpm{61.0}{1.5} \\

& screw & \resultpm{94.4}{1.5} & \resultpm{97.8}{0.8} & \resultpm{93.2}{0.2} & \resultpm{99.5}{0.0} & \resultpm{98.0}{0.2} & \resultpm{59.5}{2.5} \\

& tile & \resultpm{100.0}{0.0} & \resultpm{100.0}{0.0} & \resultpm{100.0}{0.0} & \resultpm{98.2}{0.1} & \resultpm{95.4}{0.2} & \resultpm{76.0}{0.6} \\

& toothbrush & \resultpm{100.0}{0.0} & \resultpm{100.0}{0.0} & \resultpm{100.0}{0.0} & \resultpm{99.5}{0.1} & \resultpm{97.6}{0.3} & \resultpm{69.3}{2.7} \\

& transistor & \resultpm{98.0}{1.7} & \resultpm{97.4}{2.1} & \resultpm{91.5}{4.3} & \resultpm{94.1}{1.1} & \resultpm{80.8}{2.4} & \resultpm{54.4}{3.2} \\

& wood & \resultpm{99.7}{0.1} & \resultpm{99.9}{0.0} & \resultpm{98.6}{0.5} & \resultpm{97.6}{0.2} & \resultpm{96.9}{0.1} & \resultpm{68.7}{0.4} \\

& zipper & \resultpm{99.9}{0.1} & \resultpm{100.0}{0.0} & \resultpm{99.3}{0.2} & \resultpm{99.3}{0.0} & \resultpm{97.9}{0.1} & \resultpm{69.0}{0.3} \\

\bottomrule
\end{tabular}
\end{center}

\clearpage

\begin{center}
\captionof{table}{
    Per-category anomaly detection and localization results on VisA under the
    1-, 2-, and 4-shot settings.
    Results are mean and sample standard deviation over three seeds.
    All values are percentages.
}
\label{tab:supp_per_object_visa}

\vspace{2pt}

\small
\renewcommand{\arraystretch}{0.96}%
\setlength{\tabcolsep}{5pt}%
\begin{tabular}{c l|ccc|ccc}
\toprule

\multirow{3}{*}{\textbf{Setup}}
& \multirow{3}{*}{\textbf{Category}}
& \multicolumn{6}{c}{\textbf{VisA}}
\\

&
& \multicolumn{3}{c|}{\textbf{Image-level}}
& \multicolumn{3}{c}{\textbf{Pixel-level}}
\\

&
& AUROC & AUPR & \(F_{1,\max}\)
& AUROC & AUPRO & \(F_{1,\max}\)
\\

\midrule

% =========================================================
% 1-shot
% =========================================================

\multirow{12}{*}{\textbf{1-shot}}
& candle & \resultpm{95.3}{0.6} & \resultpm{95.7}{0.4} & \resultpm{90.4}{0.7} & \resultpm{99.6}{0.0} & \resultpm{97.2}{0.2} & \resultpm{49.3}{0.9} \\

& capsules & \resultpm{98.6}{0.5} & \resultpm{99.2}{0.2} & \resultpm{95.7}{0.9} & \resultpm{99.1}{0.2} & \resultpm{98.1}{0.2} & \resultpm{61.3}{1.1} \\

& cashew & \resultpm{97.8}{0.3} & \resultpm{98.9}{0.2} & \resultpm{94.7}{0.4} & \resultpm{99.1}{0.0} & \resultpm{98.0}{0.3} & \resultpm{59.0}{0.5} \\

& chewinggum & \resultpm{99.6}{0.2} & \resultpm{99.8}{0.1} & \resultpm{97.8}{0.8} & \resultpm{99.8}{0.0} & \resultpm{96.4}{0.2} & \resultpm{73.2}{0.8} \\

& fryum & \resultpm{97.1}{0.5} & \resultpm{98.8}{0.2} & \resultpm{94.3}{1.0} & \resultpm{96.8}{0.2} & \resultpm{94.5}{0.4} & \resultpm{47.1}{0.4} \\

& macaroni1 & \resultpm{94.0}{0.9} & \resultpm{94.5}{0.9} & \resultpm{87.2}{2.1} & \resultpm{99.8}{0.0} & \resultpm{98.1}{0.4} & \resultpm{36.8}{1.1} \\

& macaroni2 & \resultpm{84.1}{7.6} & \resultpm{83.1}{8.0} & \resultpm{79.1}{6.1} & \resultpm{99.7}{0.2} & \resultpm{97.0}{1.7} & \resultpm{36.3}{1.5} \\

& pcb1 & \resultpm{95.3}{0.3} & \resultpm{94.0}{0.2} & \resultpm{92.7}{0.7} & \resultpm{99.4}{0.0} & \resultpm{95.7}{0.2} & \resultpm{74.9}{0.6} \\

& pcb2 & \resultpm{93.7}{1.3} & \resultpm{93.2}{1.2} & \resultpm{88.7}{2.0} & \resultpm{97.6}{0.0} & \resultpm{94.1}{0.2} & \resultpm{53.5}{0.1} \\

& pcb3 & \resultpm{93.1}{1.1} & \resultpm{92.6}{1.4} & \resultpm{88.1}{0.6} & \resultpm{96.6}{0.2} & \resultpm{94.0}{0.5} & \resultpm{53.7}{1.9} \\

& pcb4 & \resultpm{99.5}{0.1} & \resultpm{99.4}{0.1} & \resultpm{97.7}{0.3} & \resultpm{98.0}{0.0} & \resultpm{95.0}{0.4} & \resultpm{47.0}{0.3} \\

& pipe\_fryum & \resultpm{96.1}{0.7} & \resultpm{98.0}{0.4} & \resultpm{94.5}{1.5} & \resultpm{99.0}{0.1} & \resultpm{98.0}{0.1} & \resultpm{55.6}{3.1} \\

\midrule

% =========================================================
% 2-shot
% =========================================================

\multirow{12}{*}{\textbf{2-shot}}
& candle & \resultpm{95.3}{0.4} & \resultpm{95.4}{0.5} & \resultpm{89.8}{0.2} & \resultpm{99.7}{0.0} & \resultpm{97.6}{0.1} & \resultpm{49.8}{0.2} \\

& capsules & \resultpm{99.1}{0.4} & \resultpm{99.4}{0.3} & \resultpm{97.0}{0.5} & \resultpm{99.2}{0.2} & \resultpm{98.5}{0.1} & \resultpm{58.5}{0.9} \\

& cashew & \resultpm{98.4}{0.1} & \resultpm{99.2}{0.1} & \resultpm{95.5}{0.6} & \resultpm{99.3}{0.1} & \resultpm{98.1}{0.1} & \resultpm{61.3}{1.3} \\

& chewinggum & \resultpm{99.6}{0.1} & \resultpm{99.8}{0.0} & \resultpm{97.7}{0.3} & \resultpm{99.8}{0.0} & \resultpm{96.4}{0.2} & \resultpm{73.2}{0.6} \\

& fryum & \resultpm{98.2}{0.1} & \resultpm{99.2}{0.1} & \resultpm{95.5}{0.3} & \resultpm{97.0}{0.1} & \resultpm{94.5}{0.1} & \resultpm{48.1}{0.4} \\

& macaroni1 & \resultpm{95.6}{1.1} & \resultpm{95.5}{0.9} & \resultpm{90.5}{2.2} & \resultpm{99.8}{0.0} & \resultpm{98.2}{0.3} & \resultpm{35.5}{0.2} \\

& macaroni2 & \resultpm{87.7}{2.3} & \resultpm{84.9}{2.3} & \resultpm{83.2}{1.9} & \resultpm{99.8}{0.1} & \resultpm{97.7}{0.6} & \resultpm{37.0}{1.6} \\

& pcb1 & \resultpm{94.4}{1.0} & \resultpm{93.1}{0.6} & \resultpm{91.6}{0.8} & \resultpm{99.5}{0.0} & \resultpm{95.9}{0.2} & \resultpm{75.5}{0.6} \\

& pcb2 & \resultpm{94.2}{0.7} & \resultpm{92.5}{1.6} & \resultpm{90.0}{0.8} & \resultpm{97.7}{0.0} & \resultpm{94.1}{0.2} & \resultpm{52.9}{0.4} \\

& pcb3 & \resultpm{94.6}{1.6} & \resultpm{94.6}{1.6} & \resultpm{90.5}{1.6} & \resultpm{97.3}{0.2} & \resultpm{94.9}{0.6} & \resultpm{56.0}{2.6} \\

& pcb4 & \resultpm{99.4}{0.2} & \resultpm{99.3}{0.2} & \resultpm{97.7}{0.7} & \resultpm{98.2}{0.1} & \resultpm{94.9}{0.4} & \resultpm{46.9}{1.3} \\

& pipe\_fryum & \resultpm{96.6}{0.6} & \resultpm{98.2}{0.3} & \resultpm{94.6}{1.3} & \resultpm{99.1}{0.1} & \resultpm{97.9}{0.1} & \resultpm{56.3}{2.1} \\

\midrule

% =========================================================
% 4-shot
% =========================================================

\multirow{12}{*}{\textbf{4-shot}}
& candle & \resultpm{94.9}{1.1} & \resultpm{94.8}{1.1} & \resultpm{89.5}{2.4} & \resultpm{99.7}{0.0} & \resultpm{97.6}{0.1} & \resultpm{49.0}{1.0} \\

& capsules & \resultpm{98.8}{0.8} & \resultpm{99.3}{0.5} & \resultpm{96.9}{1.2} & \resultpm{99.3}{0.1} & \resultpm{98.5}{0.2} & \resultpm{54.0}{2.9} \\

& cashew & \resultpm{99.0}{0.3} & \resultpm{99.5}{0.1} & \resultpm{97.2}{0.7} & \resultpm{99.4}{0.0} & \resultpm{97.9}{0.2} & \resultpm{63.1}{1.0} \\

& chewinggum & \resultpm{99.5}{0.1} & \resultpm{99.8}{0.0} & \resultpm{97.6}{0.5} & \resultpm{99.8}{0.0} & \resultpm{96.3}{0.1} & \resultpm{73.2}{0.9} \\

& fryum & \resultpm{98.6}{0.4} & \resultpm{99.4}{0.2} & \resultpm{95.7}{0.7} & \resultpm{97.3}{0.1} & \resultpm{94.6}{0.1} & \resultpm{48.7}{0.2} \\

& macaroni1 & \resultpm{95.7}{1.6} & \resultpm{95.7}{1.3} & \resultpm{90.4}{3.0} & \resultpm{99.8}{0.0} & \resultpm{98.3}{0.0} & \resultpm{34.7}{1.4} \\

& macaroni2 & \resultpm{88.4}{3.7} & \resultpm{85.2}{5.2} & \resultpm{83.5}{2.7} & \resultpm{99.8}{0.0} & \resultpm{98.2}{0.2} & \resultpm{36.6}{0.5} \\

& pcb1 & \resultpm{94.4}{0.5} & \resultpm{92.4}{0.5} & \resultpm{92.0}{0.6} & \resultpm{99.5}{0.0} & \resultpm{96.2}{0.0} & \resultpm{76.1}{0.3} \\

& pcb2 & \resultpm{93.8}{1.3} & \resultpm{91.9}{2.8} & \resultpm{89.8}{0.4} & \resultpm{97.8}{0.1} & \resultpm{94.6}{0.2} & \resultpm{51.5}{1.3} \\

& pcb3 & \resultpm{95.1}{1.1} & \resultpm{94.9}{0.9} & \resultpm{90.1}{1.7} & \resultpm{97.8}{0.0} & \resultpm{95.3}{0.4} & \resultpm{56.5}{1.7} \\

& pcb4 & \resultpm{99.4}{0.1} & \resultpm{99.3}{0.1} & \resultpm{97.6}{0.5} & \resultpm{98.4}{0.2} & \resultpm{95.4}{0.1} & \resultpm{47.8}{1.8} \\

& pipe\_fryum & \resultpm{98.2}{0.1} & \resultpm{99.0}{0.1} & \resultpm{96.7}{0.7} & \resultpm{99.3}{0.0} & \resultpm{97.9}{0.1} & \resultpm{59.6}{1.3} \\

\bottomrule
\end{tabular}
\end{center}

\clearpage

\begin{center}
\captionof{table}{
    Per-category anomaly detection and localization results on Real-IAD under the
    1- and 2-sample settings. Under the sample-shot protocol, each physical
    sample comprises five fixed multi-view captures (5 and 10 support images, respectively).
    Results are mean and sample standard deviation over three seeds.
    All values are percentages.
}
\label{tab:supp_per_object_realiad_wide_1_2}

\vspace{2pt}

\footnotesize
\renewcommand{\arraystretch}{0.96}%
\setlength{\tabcolsep}{3.8pt}%
\resizebox{\textwidth}{!}{%
\begin{tabular}{l|ccc|ccc|ccc|ccc}
\toprule

\multirow{3}{*}{\textbf{Category}}
& \multicolumn{6}{c|}{\textbf{1-sample}}
& \multicolumn{6}{c}{\textbf{2-sample}}
\\

& \multicolumn{3}{c|}{\textbf{Image-level}}
& \multicolumn{3}{c|}{\textbf{Pixel-level}}
& \multicolumn{3}{c|}{\textbf{Image-level}}
& \multicolumn{3}{c}{\textbf{Pixel-level}}
\\

& AUROC & AUPR & \(F_{1,\max}\) & AUROC & AUPRO & \(F_{1,\max}\) & AUROC & AUPR & \(F_{1,\max}\) & AUROC & AUPRO & \(F_{1,\max}\)
\\

\midrule

audiojack & \resultpm{79.2}{6.7} & \resultpm{67.1}{9.7} & \resultpm{62.9}{5.8} & \resultpm{99.0}{0.1} & \resultpm{94.1}{0.4} & \resultpm{42.6}{7.7} & \resultpm{79.3}{9.8} & \resultpm{65.9}{13.7} & \resultpm{64.0}{6.3} & \resultpm{99.0}{0.1} & \resultpm{94.0}{0.5} & \resultpm{42.9}{5.9} \\

bottle\_cap & \resultpm{91.2}{0.3} & \resultpm{88.8}{0.4} & \resultpm{81.0}{0.4} & \resultpm{99.9}{0.0} & \resultpm{99.3}{0.1} & \resultpm{43.6}{0.2} & \resultpm{91.5}{0.9} & \resultpm{89.4}{1.0} & \resultpm{81.2}{0.4} & \resultpm{99.9}{0.0} & \resultpm{99.3}{0.1} & \resultpm{42.0}{0.5} \\

button\_battery & \resultpm{87.0}{0.7} & \resultpm{88.8}{0.6} & \resultpm{83.3}{0.7} & \resultpm{99.4}{0.0} & \resultpm{96.0}{0.1} & \resultpm{47.9}{1.0} & \resultpm{88.2}{1.2} & \resultpm{89.0}{0.8} & \resultpm{84.6}{1.5} & \resultpm{99.4}{0.0} & \resultpm{95.9}{0.2} & \resultpm{46.5}{0.9} \\

end\_cap & \resultpm{86.4}{2.0} & \resultpm{88.1}{0.8} & \resultpm{83.0}{2.1} & \resultpm{99.5}{0.1} & \resultpm{98.5}{0.2} & \resultpm{38.8}{1.5} & \resultpm{88.4}{0.1} & \resultpm{89.3}{0.4} & \resultpm{84.5}{0.6} & \resultpm{99.6}{0.0} & \resultpm{98.7}{0.0} & \resultpm{37.5}{1.0} \\

eraser & \resultpm{92.9}{0.9} & \resultpm{89.9}{0.9} & \resultpm{82.3}{1.5} & \resultpm{99.9}{0.0} & \resultpm{99.3}{0.0} & \resultpm{51.0}{1.1} & \resultpm{92.9}{0.6} & \resultpm{89.8}{0.6} & \resultpm{82.1}{0.7} & \resultpm{99.9}{0.0} & \resultpm{99.3}{0.0} & \resultpm{50.1}{0.7} \\

fire\_hood & \resultpm{88.0}{0.7} & \resultpm{80.5}{1.1} & \resultpm{73.8}{1.0} & \resultpm{99.8}{0.0} & \resultpm{98.4}{0.2} & \resultpm{52.5}{1.3} & \resultpm{87.6}{0.7} & \resultpm{80.0}{1.3} & \resultpm{73.6}{0.6} & \resultpm{99.8}{0.0} & \resultpm{98.3}{0.3} & \resultpm{51.8}{1.5} \\

mint & \resultpm{83.0}{1.6} & \resultpm{83.3}{1.3} & \resultpm{74.0}{1.5} & \resultpm{99.5}{0.0} & \resultpm{95.3}{0.2} & \resultpm{37.1}{0.9} & \resultpm{83.7}{1.1} & \resultpm{84.0}{0.8} & \resultpm{74.4}{1.3} & \resultpm{99.5}{0.0} & \resultpm{95.6}{0.1} & \resultpm{36.3}{0.4} \\

mounts & \resultpm{86.9}{0.2} & \resultpm{75.9}{1.1} & \resultpm{74.2}{1.1} & \resultpm{99.5}{0.0} & \resultpm{98.8}{0.0} & \resultpm{48.1}{0.8} & \resultpm{87.2}{0.2} & \resultpm{76.5}{0.9} & \resultpm{75.1}{1.0} & \resultpm{99.5}{0.0} & \resultpm{98.8}{0.0} & \resultpm{46.7}{1.2} \\

pcb & \resultpm{88.9}{0.7} & \resultpm{92.9}{0.8} & \resultpm{84.2}{0.1} & \resultpm{99.7}{0.0} & \resultpm{97.9}{0.0} & \resultpm{59.8}{1.4} & \resultpm{88.9}{0.8} & \resultpm{92.9}{0.7} & \resultpm{84.4}{0.2} & \resultpm{99.6}{0.0} & \resultpm{97.8}{0.1} & \resultpm{59.6}{1.2} \\

phone\_battery & \resultpm{90.3}{1.0} & \resultpm{86.6}{2.2} & \resultpm{79.8}{1.0} & \resultpm{99.9}{0.0} & \resultpm{99.0}{0.0} & \resultpm{67.9}{0.9} & \resultpm{89.8}{0.7} & \resultpm{84.9}{2.2} & \resultpm{79.8}{0.3} & \resultpm{99.9}{0.0} & \resultpm{98.9}{0.0} & \resultpm{65.2}{1.4} \\

plastic\_nut & \resultpm{87.0}{2.3} & \resultpm{76.4}{3.7} & \resultpm{71.4}{3.5} & \resultpm{99.9}{0.0} & \resultpm{99.2}{0.2} & \resultpm{48.0}{1.4} & \resultpm{87.6}{0.8} & \resultpm{77.1}{1.2} & \resultpm{71.7}{1.3} & \resultpm{99.9}{0.0} & \resultpm{99.3}{0.1} & \resultpm{45.8}{1.1} \\

plastic\_plug & \resultpm{84.3}{1.7} & \resultpm{75.7}{2.2} & \resultpm{71.2}{1.6} & \resultpm{98.4}{0.1} & \resultpm{93.3}{0.8} & \resultpm{41.5}{1.0} & \resultpm{83.6}{2.2} & \resultpm{71.1}{2.3} & \resultpm{71.8}{2.4} & \resultpm{97.6}{0.3} & \resultpm{90.2}{1.2} & \resultpm{16.6}{5.1} \\

porcelain\_doll & \resultpm{87.3}{3.1} & \resultpm{76.7}{5.5} & \resultpm{72.3}{2.8} & \resultpm{99.6}{0.1} & \resultpm{98.4}{0.6} & \resultpm{40.5}{3.2} & \resultpm{85.3}{2.2} & \resultpm{70.9}{4.3} & \resultpm{71.7}{1.8} & \resultpm{99.5}{0.0} & \resultpm{97.9}{0.3} & \resultpm{37.3}{1.5} \\

regulator & \resultpm{76.5}{1.6} & \resultpm{65.3}{2.3} & \resultpm{56.7}{1.0} & \resultpm{99.6}{0.0} & \resultpm{96.9}{0.1} & \resultpm{49.7}{0.7} & \resultpm{79.6}{4.8} & \resultpm{69.3}{6.2} & \resultpm{60.1}{5.8} & \resultpm{99.6}{0.0} & \resultpm{97.3}{0.1} & \resultpm{51.3}{1.9} \\

rolled\_strip\_base & \resultpm{93.5}{3.1} & \resultpm{96.2}{2.0} & \resultpm{91.2}{1.9} & \resultpm{99.9}{0.0} & \resultpm{99.3}{0.1} & \resultpm{54.4}{0.6} & \resultpm{94.3}{2.6} & \resultpm{96.6}{1.7} & \resultpm{92.2}{1.7} & \resultpm{99.8}{0.0} & \resultpm{99.2}{0.1} & \resultpm{51.2}{0.8} \\

sim\_card\_set & \resultpm{95.8}{0.2} & \resultpm{96.5}{0.1} & \resultpm{90.3}{0.6} & \resultpm{99.9}{0.0} & \resultpm{99.0}{0.1} & \resultpm{64.7}{0.6} & \resultpm{96.0}{0.2} & \resultpm{96.5}{0.1} & \resultpm{90.5}{0.5} & \resultpm{99.9}{0.0} & \resultpm{98.9}{0.1} & \resultpm{64.0}{0.3} \\

switch & \resultpm{89.1}{2.1} & \resultpm{91.3}{1.8} & \resultpm{81.0}{2.1} & \resultpm{96.3}{0.1} & \resultpm{94.8}{0.0} & \resultpm{56.8}{1.0} & \resultpm{90.3}{0.9} & \resultpm{92.2}{0.8} & \resultpm{82.3}{0.7} & \resultpm{96.6}{0.1} & \resultpm{95.1}{0.2} & \resultpm{57.0}{1.1} \\

tape & \resultpm{96.8}{0.3} & \resultpm{95.4}{0.5} & \resultpm{88.3}{1.1} & \resultpm{99.9}{0.0} & \resultpm{99.5}{0.0} & \resultpm{60.5}{0.5} & \resultpm{97.1}{0.3} & \resultpm{95.4}{0.5} & \resultpm{88.5}{0.7} & \resultpm{99.9}{0.0} & \resultpm{99.5}{0.0} & \resultpm{59.3}{1.1} \\

terminalblock & \resultpm{89.7}{0.8} & \resultpm{91.4}{0.5} & \resultpm{82.4}{1.3} & \resultpm{99.6}{0.0} & \resultpm{99.1}{0.0} & \resultpm{57.3}{0.5} & \resultpm{89.3}{0.8} & \resultpm{90.8}{0.9} & \resultpm{82.8}{0.5} & \resultpm{99.6}{0.0} & \resultpm{99.0}{0.0} & \resultpm{55.7}{0.2} \\

toothbrush & \resultpm{85.9}{1.5} & \resultpm{88.0}{2.0} & \resultpm{80.7}{1.1} & \resultpm{98.1}{0.1} & \resultpm{94.6}{0.4} & \resultpm{45.3}{1.5} & \resultpm{85.7}{3.0} & \resultpm{87.5}{4.1} & \resultpm{80.6}{1.2} & \resultpm{98.1}{0.1} & \resultpm{94.3}{0.2} & \resultpm{45.3}{2.5} \\

toy & \resultpm{80.5}{1.2} & \resultpm{85.8}{1.2} & \resultpm{77.8}{0.8} & \resultpm{94.2}{0.2} & \resultpm{91.7}{0.4} & \resultpm{37.2}{0.3} & \resultpm{83.1}{1.6} & \resultpm{88.0}{1.5} & \resultpm{79.7}{1.2} & \resultpm{93.9}{0.1} & \resultpm{91.6}{0.2} & \resultpm{37.6}{0.1} \\

toy\_brick & \resultpm{82.4}{1.5} & \resultpm{76.9}{2.8} & \resultpm{71.4}{1.4} & \resultpm{99.1}{0.0} & \resultpm{92.6}{1.4} & \resultpm{46.5}{0.6} & \resultpm{83.1}{0.1} & \resultpm{77.8}{0.9} & \resultpm{71.7}{0.1} & \resultpm{99.0}{0.1} & \resultpm{91.4}{0.6} & \resultpm{45.9}{0.8} \\

transistor1 & \resultpm{90.5}{0.6} & \resultpm{90.8}{1.0} & \resultpm{87.1}{0.4} & \resultpm{99.2}{0.1} & \resultpm{96.5}{0.4} & \resultpm{54.4}{0.9} & \resultpm{92.4}{1.7} & \resultpm{93.0}{2.5} & \resultpm{88.3}{0.4} & \resultpm{99.3}{0.0} & \resultpm{96.9}{0.2} & \resultpm{55.0}{1.5} \\

u\_block & \resultpm{85.5}{1.3} & \resultpm{71.5}{2.3} & \resultpm{71.1}{2.0} & \resultpm{99.7}{0.0} & \resultpm{98.5}{0.2} & \resultpm{37.6}{1.8} & \resultpm{87.2}{1.0} & \resultpm{74.8}{2.5} & \resultpm{72.5}{0.9} & \resultpm{99.7}{0.0} & \resultpm{98.6}{0.2} & \resultpm{39.0}{1.4} \\

usb & \resultpm{87.9}{1.2} & \resultpm{83.4}{1.7} & \resultpm{78.9}{0.9} & \resultpm{98.8}{0.1} & \resultpm{97.0}{0.2} & \resultpm{45.7}{0.8} & \resultpm{87.6}{0.2} & \resultpm{82.6}{0.3} & \resultpm{78.8}{0.4} & \resultpm{98.8}{0.1} & \resultpm{96.9}{0.1} & \resultpm{44.4}{0.2} \\

usb\_adaptor & \resultpm{88.3}{1.0} & \resultpm{84.2}{1.2} & \resultpm{75.4}{2.7} & \resultpm{99.8}{0.0} & \resultpm{98.4}{0.2} & \resultpm{42.3}{0.1} & \resultpm{88.1}{0.7} & \resultpm{83.8}{1.1} & \resultpm{74.8}{1.7} & \resultpm{99.8}{0.0} & \resultpm{98.2}{0.2} & \resultpm{40.7}{0.4} \\

vcpill & \resultpm{91.4}{2.2} & \resultpm{90.2}{2.4} & \resultpm{80.2}{2.8} & \resultpm{99.5}{0.1} & \resultpm{97.5}{0.8} & \resultpm{67.8}{0.9} & \resultpm{92.8}{1.1} & \resultpm{91.6}{1.3} & \resultpm{82.5}{1.7} & \resultpm{99.5}{0.0} & \resultpm{97.7}{0.2} & \resultpm{67.8}{0.3} \\

wooden\_beads & \resultpm{88.9}{0.8} & \resultpm{87.5}{1.1} & \resultpm{78.6}{0.7} & \resultpm{99.6}{0.0} & \resultpm{95.2}{1.6} & \resultpm{54.4}{0.9} & \resultpm{89.7}{0.5} & \resultpm{88.4}{0.9} & \resultpm{79.9}{0.8} & \resultpm{99.6}{0.0} & \resultpm{95.4}{0.9} & \resultpm{53.5}{0.8} \\

woodstick & \resultpm{90.9}{0.9} & \resultpm{82.0}{1.0} & \resultpm{74.4}{0.7} & \resultpm{99.6}{0.0} & \resultpm{96.9}{0.1} & \resultpm{60.2}{0.6} & \resultpm{91.0}{0.6} & \resultpm{81.7}{1.5} & \resultpm{74.4}{1.7} & \resultpm{99.6}{0.0} & \resultpm{96.8}{0.3} & \resultpm{59.1}{2.2} \\

zipper & \resultpm{98.7}{0.2} & \resultpm{99.1}{0.1} & \resultpm{96.0}{0.2} & \resultpm{99.5}{0.0} & \resultpm{98.7}{0.1} & \resultpm{64.7}{1.3} & \resultpm{98.4}{0.2} & \resultpm{98.9}{0.2} & \resultpm{95.7}{0.1} & \resultpm{99.5}{0.0} & \resultpm{98.6}{0.0} & \resultpm{63.8}{0.4} \\

\bottomrule
\end{tabular}%
}
\end{center}

\clearpage

\begin{center}
\captionof{table}{
    Per-category anomaly detection and localization results on Real-IAD under the
    4-sample setting. Under the sample-shot protocol, each physical
    sample comprises five fixed multi-view captures (20 support images).
    Results are mean and sample standard deviation over three seeds.
    All values are percentages.
}
\label{tab:supp_per_object_realiad_wide_4}

\vspace{2pt}

\small
\renewcommand{\arraystretch}{0.96}%
\setlength{\tabcolsep}{5pt}%
\begin{tabular}{l|ccc|ccc}
\toprule

\multirow{3}{*}{\textbf{Category}}
& \multicolumn{6}{c}{\textbf{4-sample}}
\\

& \multicolumn{3}{c|}{\textbf{Image-level}}
& \multicolumn{3}{c}{\textbf{Pixel-level}}
\\

& AUROC & AUPR & \(F_{1,\max}\) & AUROC & AUPRO & \(F_{1,\max}\)
\\

\midrule

audiojack & \resultpm{80.8}{5.5} & \resultpm{70.2}{6.6} & \resultpm{64.5}{5.3} & \resultpm{99.0}{0.0} & \resultpm{94.0}{0.1} & \resultpm{47.2}{2.1} \\

bottle\_cap & \resultpm{93.0}{0.3} & \resultpm{91.1}{0.4} & \resultpm{82.9}{0.4} & \resultpm{99.9}{0.0} & \resultpm{99.4}{0.0} & \resultpm{41.2}{0.3} \\

button\_battery & \resultpm{89.9}{0.7} & \resultpm{89.9}{0.6} & \resultpm{86.5}{1.1} & \resultpm{99.4}{0.0} & \resultpm{96.1}{0.1} & \resultpm{46.6}{1.4} \\

end\_cap & \resultpm{89.3}{0.3} & \resultpm{90.3}{0.4} & \resultpm{85.2}{0.2} & \resultpm{99.6}{0.0} & \resultpm{98.7}{0.1} & \resultpm{37.9}{0.6} \\

eraser & \resultpm{94.1}{0.6} & \resultpm{91.2}{1.5} & \resultpm{83.9}{0.8} & \resultpm{99.9}{0.0} & \resultpm{99.3}{0.0} & \resultpm{49.4}{0.9} \\

fire\_hood & \resultpm{88.8}{0.2} & \resultpm{81.9}{0.4} & \resultpm{75.0}{0.5} & \resultpm{99.8}{0.0} & \resultpm{98.4}{0.1} & \resultpm{53.5}{1.5} \\

mint & \resultpm{84.2}{0.1} & \resultpm{84.2}{0.0} & \resultpm{74.8}{0.2} & \resultpm{99.6}{0.0} & \resultpm{95.6}{0.0} & \resultpm{35.7}{0.2} \\

mounts & \resultpm{87.7}{0.5} & \resultpm{77.6}{1.2} & \resultpm{75.8}{0.8} & \resultpm{99.5}{0.0} & \resultpm{98.8}{0.0} & \resultpm{45.8}{0.8} \\

pcb & \resultpm{89.2}{0.6} & \resultpm{93.4}{0.4} & \resultpm{84.2}{0.5} & \resultpm{99.6}{0.0} & \resultpm{97.4}{0.1} & \resultpm{59.6}{0.7} \\

phone\_battery & \resultpm{90.2}{1.8} & \resultpm{84.9}{4.5} & \resultpm{80.5}{1.2} & \resultpm{99.9}{0.0} & \resultpm{98.9}{0.1} & \resultpm{63.4}{2.8} \\

plastic\_nut & \resultpm{89.0}{0.9} & \resultpm{79.5}{2.1} & \resultpm{73.3}{1.2} & \resultpm{99.9}{0.0} & \resultpm{99.3}{0.1} & \resultpm{45.5}{1.1} \\

plastic\_plug & \resultpm{83.6}{0.6} & \resultpm{70.3}{1.6} & \resultpm{71.8}{1.2} & \resultpm{97.3}{0.2} & \resultpm{89.2}{0.8} & \resultpm{14.1}{13.0} \\

porcelain\_doll & \resultpm{86.3}{1.0} & \resultpm{72.5}{2.6} & \resultpm{72.7}{1.1} & \resultpm{99.5}{0.0} & \resultpm{98.1}{0.2} & \resultpm{37.4}{1.5} \\

regulator & \resultpm{83.7}{1.7} & \resultpm{73.6}{2.2} & \resultpm{63.7}{1.6} & \resultpm{99.7}{0.0} & \resultpm{97.4}{0.2} & \resultpm{52.2}{0.5} \\

rolled\_strip\_base & \resultpm{96.4}{1.3} & \resultpm{97.9}{0.9} & \resultpm{93.5}{0.9} & \resultpm{99.8}{0.0} & \resultpm{99.2}{0.1} & \resultpm{49.8}{0.2} \\

sim\_card\_set & \resultpm{96.0}{0.4} & \resultpm{96.4}{0.3} & \resultpm{90.5}{0.8} & \resultpm{99.9}{0.0} & \resultpm{98.8}{0.1} & \resultpm{64.2}{0.5} \\

switch & \resultpm{89.2}{0.7} & \resultpm{91.6}{0.7} & \resultpm{81.0}{0.6} & \resultpm{96.5}{0.1} & \resultpm{94.8}{0.2} & \resultpm{56.7}{1.1} \\

tape & \resultpm{97.1}{0.3} & \resultpm{95.4}{0.5} & \resultpm{88.8}{0.5} & \resultpm{99.9}{0.0} & \resultpm{99.5}{0.0} & \resultpm{58.7}{0.7} \\

terminalblock & \resultpm{87.7}{0.6} & \resultpm{89.1}{0.5} & \resultpm{82.4}{0.8} & \resultpm{99.6}{0.0} & \resultpm{98.9}{0.0} & \resultpm{55.0}{0.2} \\

toothbrush & \resultpm{87.8}{0.2} & \resultpm{90.1}{0.4} & \resultpm{81.7}{0.4} & \resultpm{98.1}{0.1} & \resultpm{94.5}{0.4} & \resultpm{46.5}{0.3} \\

toy & \resultpm{84.3}{1.7} & \resultpm{89.4}{1.1} & \resultpm{80.4}{1.4} & \resultpm{93.2}{0.2} & \resultpm{90.9}{0.5} & \resultpm{37.7}{0.2} \\

toy\_brick & \resultpm{84.4}{0.6} & \resultpm{79.6}{1.7} & \resultpm{73.3}{0.5} & \resultpm{99.1}{0.1} & \resultpm{91.5}{0.9} & \resultpm{46.7}{1.6} \\

transistor1 & \resultpm{93.6}{2.1} & \resultpm{94.6}{2.6} & \resultpm{89.1}{1.9} & \resultpm{99.3}{0.1} & \resultpm{96.8}{0.2} & \resultpm{55.8}{1.1} \\

u\_block & \resultpm{88.2}{0.7} & \resultpm{75.8}{1.3} & \resultpm{74.3}{1.1} & \resultpm{99.7}{0.0} & \resultpm{98.7}{0.1} & \resultpm{38.8}{0.9} \\

usb & \resultpm{87.0}{0.7} & \resultpm{82.4}{1.0} & \resultpm{77.8}{0.5} & \resultpm{98.6}{0.1} & \resultpm{96.4}{0.2} & \resultpm{44.0}{0.9} \\

usb\_adaptor & \resultpm{88.3}{0.6} & \resultpm{84.1}{0.9} & \resultpm{75.2}{1.0} & \resultpm{99.7}{0.0} & \resultpm{97.9}{0.3} & \resultpm{40.5}{0.4} \\

vcpill & \resultpm{93.8}{0.3} & \resultpm{92.6}{0.5} & \resultpm{83.7}{0.5} & \resultpm{99.6}{0.0} & \resultpm{97.9}{0.3} & \resultpm{67.9}{0.6} \\

wooden\_beads & \resultpm{90.2}{0.5} & \resultpm{88.9}{0.7} & \resultpm{80.3}{0.8} & \resultpm{99.7}{0.0} & \resultpm{96.1}{0.7} & \resultpm{53.7}{0.4} \\

woodstick & \resultpm{91.7}{0.8} & \resultpm{82.9}{1.6} & \resultpm{75.4}{1.5} & \resultpm{99.7}{0.0} & \resultpm{97.0}{0.2} & \resultpm{60.2}{1.3} \\

zipper & \resultpm{98.7}{0.1} & \resultpm{99.1}{0.1} & \resultpm{96.3}{0.0} & \resultpm{99.5}{0.0} & \resultpm{98.6}{0.0} & \resultpm{63.6}{0.4} \\

\bottomrule
\end{tabular}
\end{center}

\FloatBarrier

\end{document}